\documentclass[manuscript,screen,nonacm]{acmart}

\AtBeginDocument{%
  }

\usepackage{booktabs}
\usepackage{tabularx}
\usepackage{graphicx}
\usepackage{multirow}
\usepackage{makecell}
\usepackage{caption} 
\usepackage{wrapfig}
\usepackage{soul}    

\usepackage{adjustbox}

\begin{document}

\title{ApneaEye : Contactless Sleep Apnea Monitoring by Measuring Nasal Respiration and Thoracoabdominal Effort using Commodity Thermal Camera}



\author{Ayush Shrivastava}
\email{shrivastavaayush@iitgn.ac.in}
\orcid{0009-0003-7241-0447}
\affiliation{
  \institution{IIT Gandhinagar}
  \city{Gandhinagar}
  \state{Gujarat}
  \country{India}
}

\author{Parv Thacker}
\email{parv.thacker@iitgn.ac.in}
\orcid{0009-0008-4813-6640}
\affiliation{%
  \institution{IIT Gandhinagar}
  \city{Gandhinagar}
  \state{Gujarat}
  \country{India}
}

\author{Rishiraj Adhikary}
\orcid{0000-0001-8089-9498}
\email{rishiraj.inbox@gmail.com}
\affiliation{%
  \institution{University of Cambridge}
  \city{Cambridge}
  \country{England}
}

\author{Saksham Bhutani}
\email{sbhutani@andrew.cmu.edu}
\orcid{0000-0001-9349-3120}
\affiliation{%
 \institution{Carnegie Mellon University}
 \city{Pittsburgh}
 \state{Pennsylvania}
 \country{USA}
 }

\author{Saurabh Mittal}
\orcid{0000-0002-7979-6405}
\email{saurabh\_kgmu@yahoo.co.in}
\affiliation{%
  \institution{All India Institute of Medical Science}
  \city{New Delhi}
  \state{Dehi}
  \country{India}
  }

\author{Mayank Goel}
\email{mayankgoel@cmu.edu}
\orcid{0000-0003-1237-7545}
\affiliation{%
 \institution{Carnegie Mellon University}
 \city{Pittsburgh}
 \state{Pennsylvania}
 \country{USA}
 }

\author{Nipun Batra}
\orcid{0000-0002-0736-7169}
\email{nipun.batra@iitgn.ac.in}
\affiliation{%
  \institution{IIT Gandhinagar}
  \city{Gandhinagar}
  \state{Gujarat}
  \country{India}
}


%
\renewcommand{\shortauthors}{Shrivastava et al.}

%
\begin{abstract}
Sleep apnea affects nearly one billion people worldwide, but is widely underdiagnosed, in large part because the diagnosis is based on polysomnography (PSG), an overnight study that requires multiple contact sensors, specialized facilities, and trained technicians. 
The clinical assessment depends on three complementary signals: nasal airflow, respiratory effort of the thoracoabdominal tract, and blood oxygen saturation (SpO$_2$). Existing non-contact approaches, including WiFi, radar, acoustic, and accelerometer-based systems, capture respiratory activity primarily through body motion and cannot directly measure nasal airflow.
This matters because thoracoabdominal motion alone cannot detect a respiratory event, where breathing effort persists despite airflow cessation, from normal breathing.
Nasal airflow is therefore central to the AASM criteria for scoring apnea and hypopnea events. 
We introduce ApneaEye, a contactless system that uses a single commodity thermal camera to simultaneously capture nasal airflow and thoracoabdominal motion without any body-attached sensors.  
Temperature variations around the nostrils during inhalation and exhalation provide a direct thermal correlate of airflow, while chest motion reflects respiratory effort.
Together, these provide the two respiratory signals that clinicians rely on. 
We evaluate ApneaEye on a dataset we collected with more than 2,000 hours of synchronized overnight thermal and PSG recordings from 261 sleep clinic patients.
Adding nasal airflow to the thoracoabdominal motion improves the exact severity classification by 8.7 percentage points, with 93.3\% of nights classified within one severity band of the PSG reference. ApneaEye estimates AHI with MAE 9.7 events/h compared to 11.5 events/h when using only motion data.    
We found that adding oximetry information along with ApneaEye reduces AHI MAE to 6.8 events/h, matching the 6.9 events/h achieved with PSG.
These results demonstrate that a single thermal camera can recover the clinically relevant respiratory signals used in sleep apnea assessment, and suggest a potential path toward non-contact home sleep testing that substantially reduces reliance on body-worn sensors.
\end{abstract}





\maketitle

\section{Introduction}

Sleep apnea is a highly prevalent yet underdiagnosed respiratory disorder associated with serious cardiovascular, metabolic, and cognitive consequences~\cite{PARISH2009563}. A global analysis published in 2019 estimated that nearly one billion adults worldwide have obstructive sleep apnea (OSA), with a substantial proportion of this burden concentrated in resource-limited settings~\cite{benjafield2019estimation}. Despite its prevalence, nearly 80\% of individuals with sleep apnea remain undiagnosed~\cite{SIA2017_OSAAWareness}. A central barrier to a broader diagnosis is polysomnography (PSG), the clinical gold standard, which requires specialized equipment, multiple physiological sensors, and access to dedicated sleep laboratories~\cite{korompili2021psg}. These constraints limit its reach and motivate approaches that can be deployed outside traditional clinical settings.

\begin{figure}
    \centering
    \includegraphics[width=1.0\linewidth]{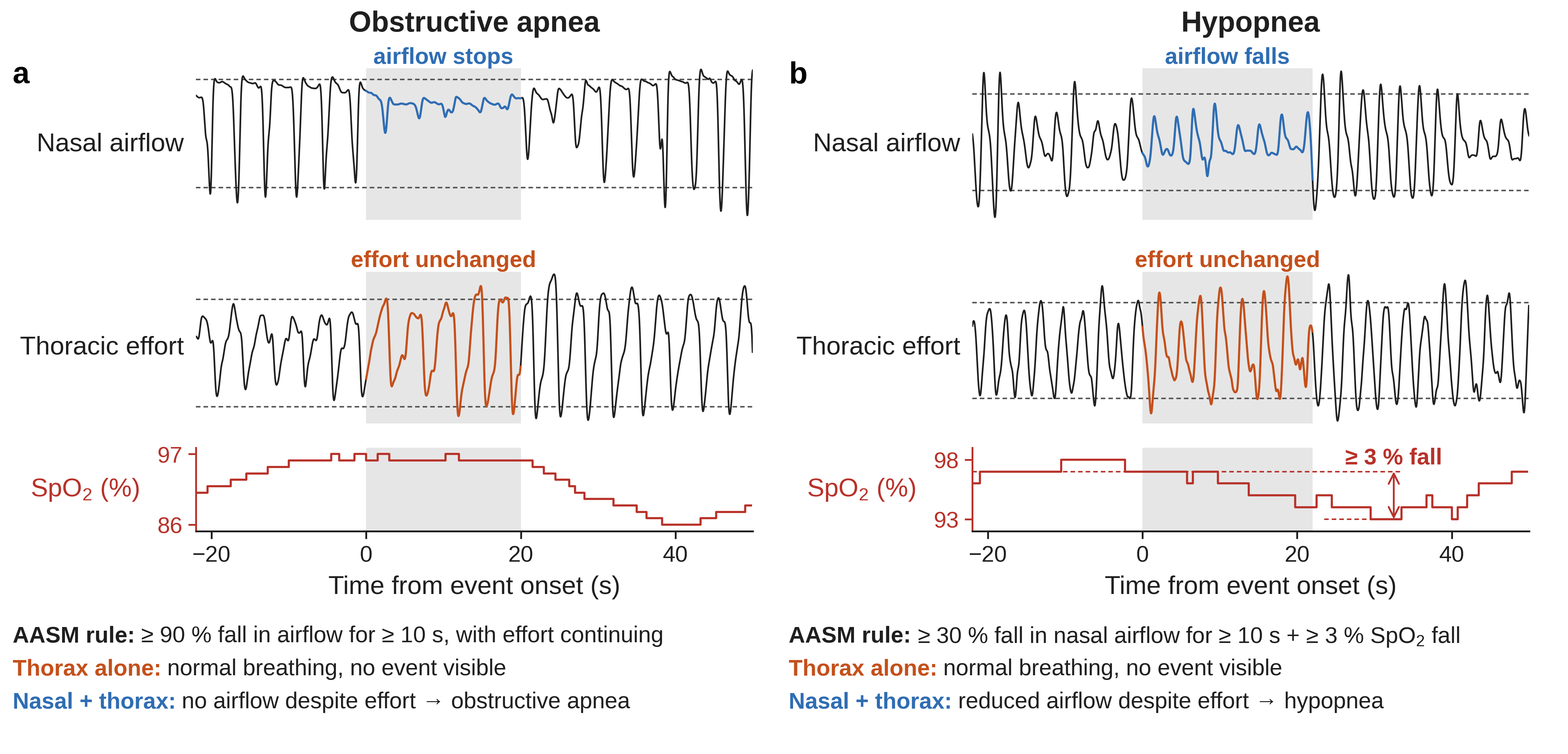}
\caption{Complementary roles of nasal airflow and thoracic effort in identifying sleep-disordered breathing events. Actual PSG signals illustrating (a) obstructive apnea, characterized by absent airflow despite continued thoracic effort, and (b) hypopnea, characterized by reduced airflow and a $\ge$ 3\% SpO$_2$ reduction despite continued effort. In both cases, thoracic effort alone does not clearly reveal the event, whereas combining nasal airflow and thoracic effort enables event identification.}
    \label{fig:Nasal_Imp}
\end{figure}

Clinical assessment of sleep apnea depends on three physiological signals: nasal airflow, thoracoabdominal respiratory effort, and blood oxygen saturation (SpO$_2$). According to the American Academy of Sleep Medicine (AASM) guidelines, nasal airflow and respiratory effort together enable the identification and classification of apnea and hypopnea events, while SpO$_2$ provides additional criteria for scoring hypopnea~\cite{berry2012aasm,berry2014aasm,berry2017aasm}.
Nasal airflow is particularly important for identifying obstructive events, as thoracoabdominal motion alone cannot reliably distinguish between normal breathing and obstructive events.
In obstructive events, the respiratory effort persists despite a substantial reduction or complete cessation of airflow. As illustrated in Figure~\ref{fig:Nasal_Imp}, both obstructive apnea and hypopnea can occur alongside continued thoracic effort, making these events difficult to identify from motion signals alone.

A growing body of work has explored wearable and non-contact approaches to sleep apnea monitoring using modalities such as Wi-Fi, radar, SONAR, cameras, and smartwatches~\cite{GuoBreathBandWifi,song2026contactless,AdhikariJoulesEye,akbarian2021_JMIR_NIR,nandakumar2015contactless,maurya2025contactless,choi2026non}. 
These approaches capture respiratory activity primarily through thoracoabdominal (thorac) motion or peripheral physiological signals such as SpO$_2$ and photoplethysmography (PPG). 
None of the approaches directly capture nasal airflow.
This gap matters because, without airflow information, these systems cannot apply the same event-scoring strategy that clinicians use, limiting both their accuracy and their interpretability in clinical contexts.

\begin{figure}  
    \centering
    \includegraphics[width=0.95\linewidth]{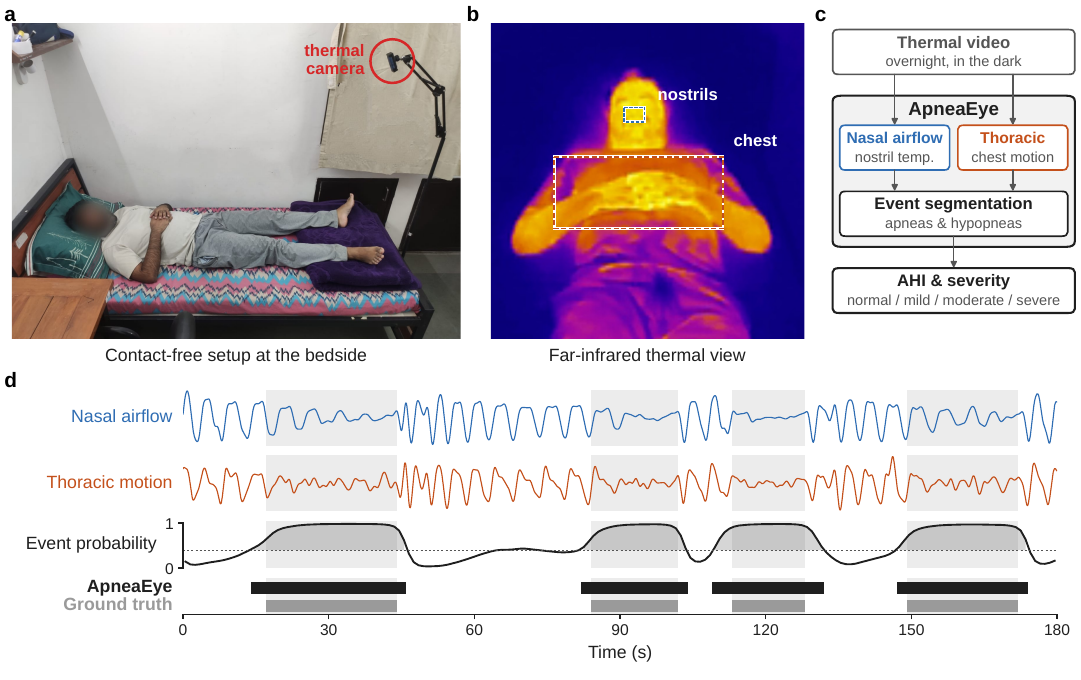}
    \caption{Overview of ApneaEye, a contact-free system for sleep apnea monitoring using far-infrared thermal imaging. (a) Bedside setup with a thermal camera recording the participant overnight. (b) Thermal view highlighting the nostril and chest regions used to extract nasal airflow and thoracic motion, respectively. (c) ApneaEye pipeline, which combines both signals to estimate per-second respiratory event probabilities and derive the apnea--hypopnea index (AHI). (d) Example 3-min recording showing nasal airflow, thoracic motion, predicted event probabilities, and ground-truth event annotations. Shaded regions indicate respiratory events.}
    \label{fig:HeroFigure}
\end{figure}

In this work, we present \textit{ApneaEye}, a contactless sleep apnea monitoring system that uses a single commodity thermal camera to simultaneously capture nasal airflow and thorac motion without any body-attached sensors~(Figure~\ref{fig:HeroFigure}).
Temperature variations around the nostrils during inhalation (cooling due to evaporation) and exhalation (difference between body and ambient temperature) provide a direct correlate of airflow that is analogous to the nasal thermistor used in PSG. 
In addition, like other camera-based approaches, ApneaEye captures the motion of the chest and abdomen to estimate the respiratory effort. 
Together, these signals allow ApneaEye to detect and characterize respiratory events using the same complementary physiological information that underlies the clinical scoring informed by the AASM guidelines. 

To build and evaluate ApneaEye, we collected a custom dataset with 2,007 hours of synchronized PSG and infrared thermal recordings from 261 patients, with all respiratory events annotated by trained sleep physicians following AASM guidelines.
To our knowledge, ApneaEye is the first system to simultaneously capture nasal airflow and thoracoabdominal motion for sleep apnea monitoring; and presents the largest clinical evaluation for such single point sensing. 
This dataset enables an evaluation against real overnight sleep on a scale that highlights robustness challenges that controlled studies cannot.

While thermal sensing of nasal airflow has been demonstrated in controlled settings for a different task of exercise detection and caloric expenditure~\cite{AdhikariJoulesEye}, making it work on extended, naturally occurring overnight clinical recordings presents substantial challenges that prior work has not addressed. Sleeping patients change positions, cover their faces, and move unpredictably.
In our dataset, the nostrils were visible during 75\% of total sleep time, dropping to around 55\% during lateral sleep. 
At a commodity thermal camera’s resolution, the nostril region spans roughly 9×6 pixels and contains background, skin edges, and motion artifacts, alongside the airflow signal. 
ApneaEye addresses these challenges in three ways. 
First, a two-stage hierarchical detector with temporal stabilization locates and tracks the nostril region reliably throughout the night. 
Second, a convolutional readout trained against synchronized PSG nasal pressure estimates respiration-related amplitude changes from noisy, low-resolution thermal patches.
Third, a per-second validity channel signals to the downstream segmentation model when nasal airflow is unavailable, allowing it to fall back to thorac motion as needed.

The primary contribution of ApneaEye is to enhance thoracoabdominal motion with nasal airflow. 
Adding nasal airflow improves the accuracy of the exact severity classification by 8.7 percentage points, with 93.3\% of patients classified within a severity band of the PSG reference.
ApneaEye's AHI estimation (MAE of 9.7 events/h) improves on motion-only estimation (11.5 events/h) and approaches the performance of a PSG system using thoracoabdominal contact sensors alone (9.6 events/h), despite relying entirely on non-contact thermal imaging.
Our experiments show that augmenting ApneaEye with oximetry reduces AHI MAE to 6.8 events/h, comparable to 6.9 events / h obtained using the corresponding PSG-based configuration. 
Importantly, these improvements arise despite nasal airflow being observable for only part of the night, demonstrating that intermittent access to a strong physiologically complementary signal can improve the evaluation of respiratory events beyond what continuous motion detection alone provides.
These findings demonstrate that a single commodity thermal camera can recover the clinically-relevant respiratory signals used in sleep apnea assessment, and suggest a practical path toward contactless home sleep testing that substantially reduces reliance on body-worn sensors. 

In summary, this paper makes the following contributions:
\begin{enumerate}
    \item ApneaEye, a contactless sleep apnea monitoring system that uses a single commodity thermal camera to simultaneously capture nasal airflow and thoracoabdominal motion without any on-body sensors.
    \item Solutions for robust nasal airflow extraction in real overnight sleep, including a two-stage hierarchical nostril detector with temporal stabilization, a convolutional readout trained against simultaneous PSG nasal pressure to isolate airflow from a noisy low-resolution pixel patch, and a per-second validity channel that allows the system to gracefully fall back to thoracoabdominal motion when nasal airflow is unavailable.
    \item A large-scale clinical evaluation comprising synchronized overnight thermal and PSG recordings of 261 patients, with physician-annotated respiratory events — a more than seven times increase from the most relevant non-contact~\cite{nandakumar2015contactless} and thermal imaging~\cite{scebba2020multispectral} prior works.
    \item Quantification of the nasal airflow contribution by showing that augmenting thoracoabdominal motion with thermal nasal airflow significantly improves apnea detection. 
\end{enumerate}

\section{Background}

\subsection{Clinical Definition of Sleep Apnea}

Sleep apnea is characterized by recurrent episodes of reduced or absent breathing during sleep, classified as apnea or hypopnea according to guidelines established by the \textit{American Academy of Sleep Medicine (AASM)}~\cite{berry2012aasm,berry2014aasm,berry2017aasm}. As summarized in Table~\ref{tab:aasm-rules}, an apnea event requires a reduction in nasal airflow of at least 90\% lasting for at least 10 seconds, while a hypopnea requires a reduction in respiratory amplitude of at least 30\% accompanied by either a 3\% drop in oxygen saturation or a cortical arousal. 
The combination of nasal airflow and thoracoabdominal effort allows clinicians to distinguish obstructive apnea, in which airflow ceases despite continued respiratory effort, from central apnea, in which both airflow and respiratory effort are absent.
The frequency of these events during sleep is quantified using the Apnea–Hypopnea Index (AHI), defined as the number of apnea and hypopnea events per hour of sleep. AHI is commonly used to assess disease severity, categorized as Normal ($AHI <5$), Mild ($5 \leq AHI < 15$), Moderate ($15 \leq AHI < 30$), and Severe ($AHI \geq 30$).

\begin{table}
\caption{Respiratory scoring rules for apnea, hypopnea, and apnea types as defined by the AASM~\cite{berry2017aasm}. These rules emphasize the importance of sensing nasal airflow together with thoracoabdominal respiratory effort.}
\label{tab:aasm-rules}
\begin{tabular}{@{}ll@{}}
\toprule
\textbf{Event} & \textbf{Diagnostic Requirement as per AASM~\cite{berry2012aasm}} \\ 

\midrule

\textbf{Hypopnea rule} &
\begin{tabular}[c]{@{}l@{}}
1. Peak signal excursions of \textit{nasal airflow} decrease by $\ge30\%$ of the pre-event baseline. \\  
2. The duration of the $\ge30\%$ drop in airflow signal lasts at least 10 seconds. \\  
3. Associated with either a $\geq3\%$ drop in \textit{blood oxygen saturation} or a cortical arousal.
\end{tabular} \\

\midrule

\textbf{Apnea rule} &
\begin{tabular}[c]{@{}l@{}}
1. A drop in the peak signal excursion by $\ge90\%$ of pre-event baseline using an \textit{airflow sensor}. \\ 
2. The duration of the $\ge90\%$ drop in airflow signal lasts at least 10 seconds.
\end{tabular} \\

\midrule

\begin{tabular}[c]{@{}l@{}}\textbf{Obstructive Sleep}\\\textbf{Apnea Event}\end{tabular} &
\begin{tabular}[c]{@{}l@{}}
1. Apnea event identified using nasal airflow (Apnea rule). \\ 
2. Presence of \textit{respiratory effort} measured using thoracic or abdominal belt signals.
\end{tabular} \\

\midrule

\begin{tabular}[c]{@{}l@{}}\textbf{Central Sleep}\\\textbf{Apnea Event}\end{tabular} &
\begin{tabular}[c]{@{}l@{}}
Apnea event where both \textit{nasal airflow} and \textit{respiratory effort} are absent.
\end{tabular} \\

\bottomrule
\end{tabular}
\end{table}

\subsection{Burden of OSA and Availability of Sleep Technicians}

Obstructive sleep apnea (OSA) represents a substantial global health burden, affecting an estimated 936 million adults worldwide, including 425 million with moderate-to-severe disease \cite{benjafield2019estimation}. However, the capacity to diagnose this large population remains constrained by the availability of sleep laboratories, specialized equipment, and trained sleep technologists. 
In the United States, a 2021 workforce survey by the American Association of Sleep Technologists (AAST) reported significant staffing challenges in 43\% of surveyed sleep laboratories~\cite{aast2021workforce}. 
The situation is particularly acute in resource-limited settings such as India, where approximately 100 million adults are estimated to have OSA, yet only around 250 dedicated sleep laboratory beds are available nationwide~\cite{garg2026sleep}. 
Existing portable and wearable alternatives often require multiple body-attached sensors, potentially compromising comfort and patient compliance. Comfortable, non-contact sensing approaches could help address these challenges by enabling less intrusive respiratory monitoring outside conventional sleep laboratories.

\begin{figure}
    \centering
        \includegraphics[width=0.90\linewidth]{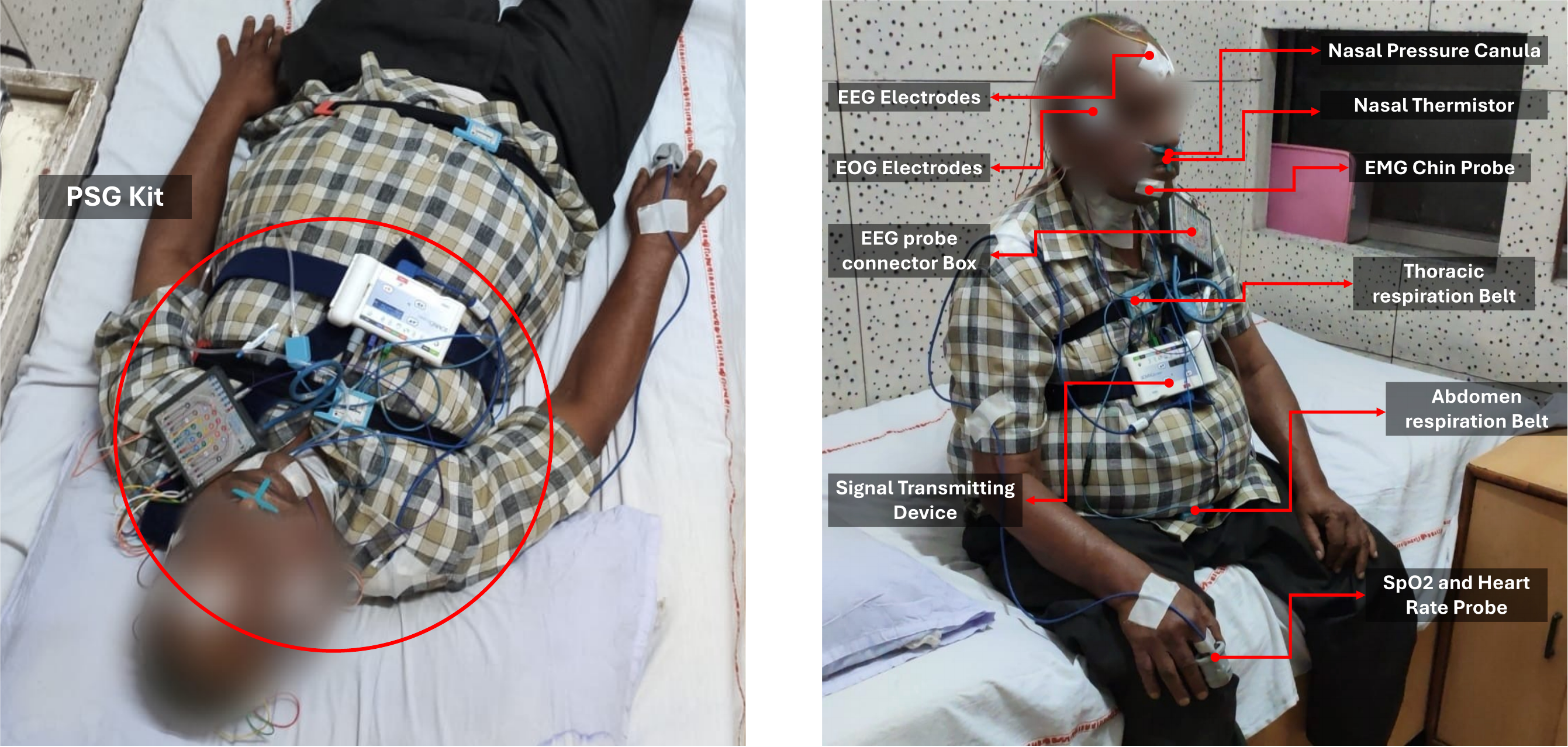}
        \caption{Apnea patient at a sleep lab wearing sensors during an
    overnight polysomnography (PSG) test.}
    \label{fig:PSG}
\end{figure}

\subsection{Polysomnography : Gold Standard for Sleep Apnea Diagnosis}


Polysomnography (PSG) is the clinical gold standard for diagnosing
sleep disorders, including sleep apnea, and records multiple
physiological signals during sleep. These signals include
electroencephalography (EEG), electrooculography (EOG),
electromyography (EMG), nasal airflow, thoracoabdominal respiratory
effort, and blood oxygen saturation (SpO$_2$). 
As shown in Figure~\ref{fig:PSG}), patients are connected to multiple wired sensors attached to the face, chest, abdomen, and fingers, requiring trained technicians for attachment and overnight monitoring. 
These logistical requirements limit the scalability and accessibility of PSG beyond specialized laboratories or expensive at-home services.

\section{Related Work}\label{sec:relatedworks}
Several non-invasive approaches have been proposed for home-based sleep apnea monitoring to address the cost and inconvenience of polysomnography (PSG). However, existing approaches differ in the physiological signals they capture and in how they identify and count respiratory events. 

\subsection{\textbf{Wearable Devices}}

Smartwatches and fitness trackers estimate sleep apnea risk using signals such as blood oxygen levels, heart rate, and body movement~\cite{chen2021apneadetector, DeepApnea_Liu, LydenSmartwatchSpO2, ArleneSmartwatchSp02_2}. These signals can indicate possible breathing problems, but they do not directly measure nasal airflow and may not capture breathing effort as clinical respiratory sensors do. As a result, these approaches often use predictions over fixed time windows or changes in other physiological signals to identify apnea events. Counting positive windows to estimate the apnea--hypopnea index (AHI) can lead to errors when multiple events occur within one window or a single event spans multiple windows.


\subsection{\textbf{Contactless Sensing Approaches}}

Several contactless approaches have been explored for respiratory monitoring using radio frequency (RF), SONAR, and Wi-Fi sensing, as well as ambient sensors such as accelerometers, seismographs, smartphones, and smart mattresses~\cite{nandakumar2015contactless, song2026contactless, song2026attention, WanMultiUserAcoustic, GuoBreathBandWifi, LiuWifiReflected}. These methods capture respiration-induced body movements or environmental changes and, in some cases, can also monitor cardiac activity through subtle heartbeat-induced movements.

Although these approaches can estimate breathing rate and detect breathing interruptions, methods that primarily rely on chest movements cannot directly capture nasal airflow. This distinction is important for sleep apnea assessment because airflow and respiratory effort may become decoupled during obstructive apnea, when breathing efforts persist despite a blocked airway. Thus, capturing both nasal airflow and thoracoabdominal motion provides complementary information for characterizing respiratory events. 

The closest work in this category is ApneaApp~\cite{nandakumar2015contactless}, which transforms a phone into an active sonar using FMCW transmissions to track thorac movements and detect apnea events. Evaluated on 37 patients in a sleep study, it demonstrated that contactless motion sensing can estimate event counts with strong correlation to PSG. However, sonar-based chest motion sensing faces the same fundamental limitation as other motion-only approaches: it cannot capture nasal airflow. 
Nandakumar~\textit{et~al.} acknowledged this directly, noting that obstructive events "\textit{might result in less noticeable changes in the breathing effort/movements but would be noticeable at the nasal pressure transducer signal}"~\cite{nandakumar2015contactless}. 
ApneaEye is designed to directly capture this missing signal by measuring temperature changes around nostrils.

\subsection{\textbf{Camera-Based Approaches}}
Cameras offer another approach to contactless sleep monitoring without requiring sensors to be attached to the body. Previous studies have explored near-infrared (NIR) cameras for sleep monitoring and apnea detection~\cite{Akbarian2020_JMIR_NIR, akbarian2021_JMIR_NIR, scebba2020multispectral, chiuNIR_2024video, choi2026non, choi2024poster}. These cameras can capture subtle thoracic movements associated with breathing but cannot directly capture the temperature variations around the nostrils caused by inhalation and exhalation. Consequently, apnea detection using conventional NIR cameras relies primarily on thoracoabdominal motion. The closest prior thermal imaging work, Scebba~\textit{et~al.}~\cite{scebba2020multispectral}, demonstrated that combining thermal nasal airflow with motion signals from NIR and FIR cameras improves respiratory rate estimation over motion alone. ApneaEye builds on this finding but addresses the substantial gap between controlled feasibility and clinical evaluation. Scebba~\textit{et~al.} evaluated 30 healthy volunteers performing simulated apnea events in a controlled supine setting, using two synchronized cameras with spatial registration. 
More recently, Choi~\textit{et~al.}\cite{choi2026non} demonstrated NIR-based obstructive sleep apnea detection across a larger clinical cohort, but like prior NIR approaches, it relies on thoracoabdominal motion and does not directly measure nasal airflow.
ApneaEye uses a single commodity thermal camera and is evaluated on 261 clinical patients with naturally occurring apnea events across real overnight sleep. It also targets a clinically different output: AHI estimation and four-class severity classification rather than respiratory rate.

\begin{table}[t]
\centering
\caption{Comparison of sensing modalities for sleep apnea monitoring. AASM guidelines require nasal airflow and thoracoabdominal effort. Only thermal imaging enables non-contact measurement of both respiratory signals.}
\label{tab:modalities}
\renewcommand{\arraystretch}{1.1}
\resizebox{\textwidth}{!}{
\begin{tabular}{lcccl}
\toprule
\textbf{Modality} & \makecell{\textbf{Nasal}\\ \textbf{Airflow}} & \makecell{\textbf{Respiratory}\\\textbf{Effort}} & \textbf{Contact} & \textbf{Comfort} \\
\midrule

Polysomnography (PSG) \cite{berry2012aasm}
& \checkmark & \checkmark & Yes 
& Low (multiple wired sensors) \\

Home Sleep Apnea Testing (HSAT)~\cite{kapur2017clinical} 
& \checkmark & \checkmark & Yes 
& Moderate (fewer sensors but still worn) \\

Smartwatches / Wearable \cite{chen2021apneadetector,LydenSmartwatchSpO2,ArleneSmartwatchSp02_2,DeepApnea_Liu}
& $\times$ & $\times$ & Yes 
& Moderate to High \\

Inertial Measurement Unit (IMU) ~\cite{hayano2025detection,boiko2023accelerometer,zhou2025accelerometer} & $\times$ & \checkmark & Yes / No$^{*}$ & Moderate to High (method dependent)  \\

WiFi-based Sensing~\cite{GuoBreathBandWifi,LiuWifiReflected,LiuWifiReflected_OffTheShelf,XieWifiCSIcirclemodel,Abdel_UBIbreath}
& $\times$ & \checkmark & No 
& High \\

Acoustic / SONAR Methods \cite{nandakumar2015contactless,Jin_ApneaDetectSnoring,DING_ApneaDetectSnoring_VGG19-LSTM,WanMultiUserAcoustic}
& $\times$ & \checkmark & No 
& High \\

NIR Camera Methods \cite{SunRespMonitorRGB,Akbarian2020_JMIR_NIR,akbarian2021_JMIR_NIR,RespError2,DengComplexMachinery,chiuNIR_2024video}
& $\times$ & \checkmark & No 
& High \\

ApneaEye via IR Imaging (Proposed)
& \checkmark & \checkmark & No 
& High \\

\bottomrule
\end{tabular}
}
\parbox{\linewidth}{\footnotesize \textit{Note:} $^{*}$ IMU-based methods can be wearable (contact) or placed in the environment (bed/mattress-mounted accelerometers) depending on the sensing setup.}

\end{table}

\subsection{\textbf{Window-Based Apnea Estimation}}
\begin{figure} [t]
    \centering
    \includegraphics[width=1.0\linewidth]{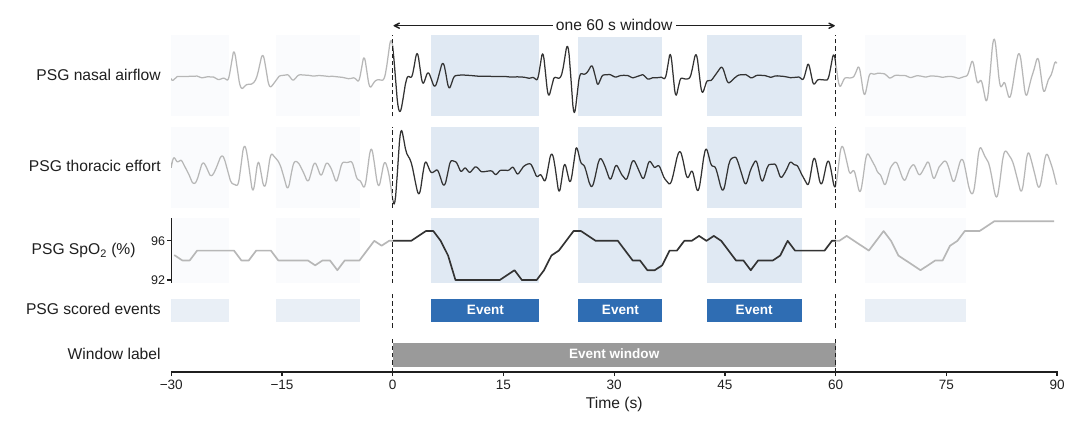}
    \caption{One 60~s window can contain several respiratory events. PSG nasal airflow (nasal pressure transducer), thoracic effort (summed respiratory inductance plethysmography) and SpO$_2$ from a test night, with the events scored and reviewed by a pair of sleep physicians. The 60~s window between the dashed lines holds three scored events. A window-level classifier labels this window once as an event window, counting one event where segmentation counts three. Airflow and effort are band-passed (0.1--1~Hz) and scaled for display.}
    \label{fig:window_problem}
\end{figure}

Systems that infer apnea burden from window-level predictions face challenges in accurately estimating the number of individual events. Window-based apnea estimation introduces a trade-off between temporal context and event-count accuracy. As discussed by Kristiansen et al.~\cite{KristiPortablePSG}, short classification windows can split a single respiratory event across multiple windows, potentially inflating the estimated event count, whereas longer windows may contain multiple events that are counted as a single positive window, leading to underestimation. 
We show this as an example in figure~\ref{fig:window_problem}.
Methods that infer event counts directly from binary window labels can suffer from splitting and merging errors, motivating approaches that explicitly segment individual respiratory events.

\subsection{Thermal Sensing for Respiration}
Long-wave infrared (LWIR) thermal imaging offers a promising approach to contactless respiratory monitoring by capturing temperature variations associated with nasal airflow and thermal patterns associated with thoracoabdominal motion. Unlike conventional cameras, thermal cameras measure emitted infrared radiation rather than reflected light, allowing them to operate independently of ambient illumination and making them particularly suitable for overnight monitoring in low-light environments. A single LWIR camera can therefore capture both respiratory signals without requiring multiple cameras or sensors attached to the body. Table~\ref{tab:modalities} summarizes the sensing capabilities and limitations of different modalities. To the best of our knowledge, no prior non-contact sleep monitoring system has simultaneously captured nasal airflow and thoracoabdominal respiratory effort using a single thermal camera and evaluated their utility for apnea assessment during real overnight sleep against PSG references.

\subsubsection{\textbf{Thermal Airflow Sensing :}}

\begin{figure}
    \centering
    \includegraphics[width=0.7\linewidth]{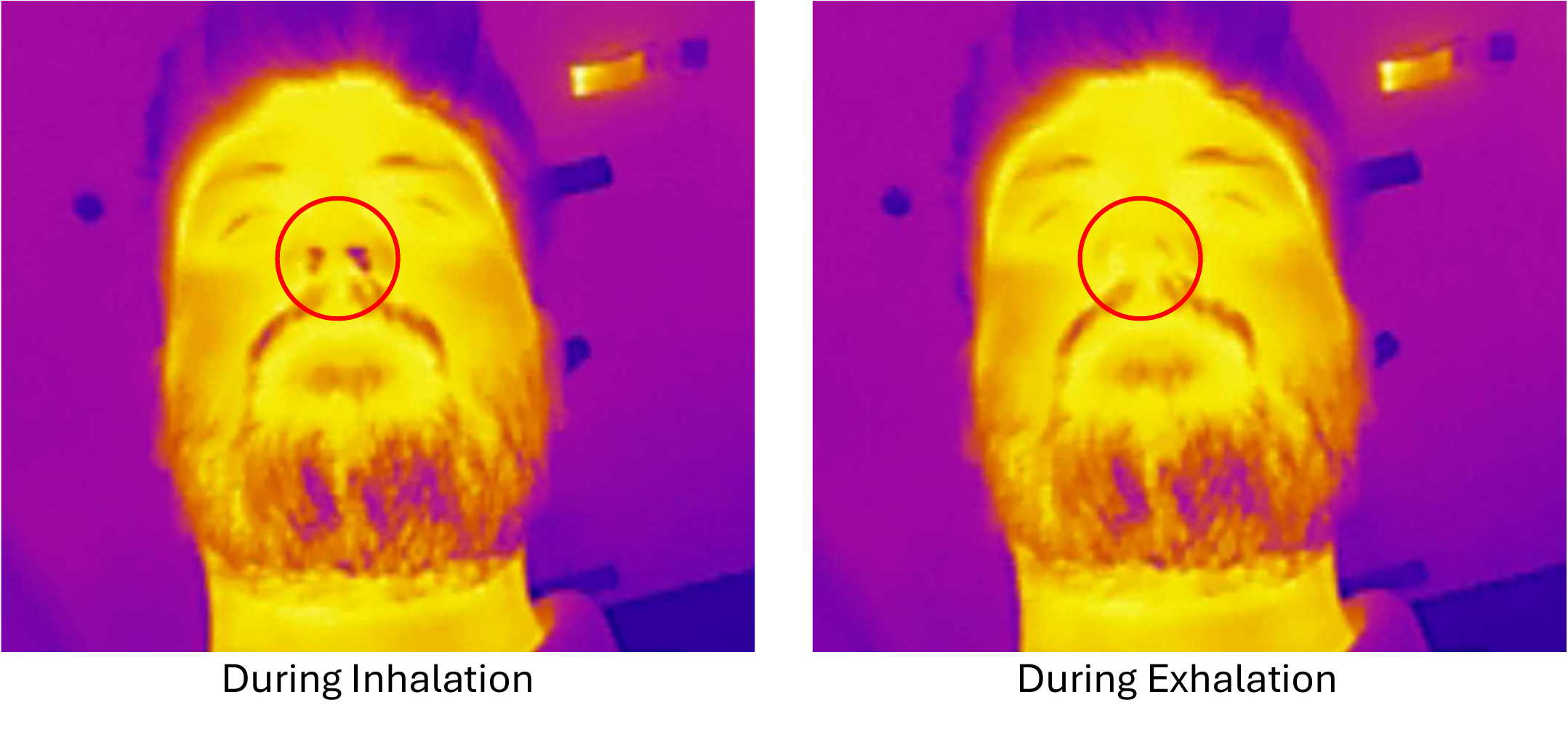}
    \caption{Thermal images showing temperature variations around the nostrils during the breathing cycle. Inhalation introduces cooler ambient air, lowering the temperature near the nostrils, while exhalation releases warmer air that appears as a thermal plume.}
    \label{fig:InhaleExhale}
\end{figure}

Breathing produces periodic temperature variations around the nostrils that are observable in infrared thermal imagery. During inhalation, cooler ambient air enters the nasal cavity and lowers the temperature near the nostrils. During exhalation, warmer air from the respiratory tract increases the local temperature, creating a visible thermal plume. This phenomenon is shown in Figure~\ref{fig:InhaleExhale}.
These alternating temperature changes follow the breathing cycle and provide a measurable indicator of nasal airflow~\cite{RespError4,RespError5,AdhikariJoulesEye}. The resulting thermal signal provides a non-contact correlate of nasal airflow, which is conventionally measured using a thermistor in PSG.

\subsubsection{\textbf{Thermal Thoracoabdominal Motion Sensing :}}
Respiration also generates periodic expansion and contraction of the chest and abdomen, reflecting respiratory effort. In thermal imagery, these movements appear as subtle displacement of heat patterns on the torso. Temporal changes in the thoracoabdominal region can therefore be analyzed to estimate respiratory motion, providing a proxy for the effort signals measured by respiratory belts in clinical polysomnography~\cite{chiuNIR_2024video,Akbarian2020_JMIR_NIR,akbarian2021_JMIR_NIR}.

\section{ApneaEye System}\label{sec:approach}

ApneaEye detects sleep apnea events from overnight thermal recordings captured using a long-wave infrared (LWIR) thermal camera. From the captured videos, it extracts thermal signals representing nasal airflow and thoracoabdominal motion, which are processed by an event-segmentation model to identify individual events. The detected events are counted to estimate the Apnea-Hypopnea Index (AHI), defined as the number of apnea and hypopnea events per hour of sleep. We evaluate ApneaEye by comparing its estimates against simultaneously acquired overnight polysomnography (PSG) recordings, which serve as the reference for respiratory event annotations and AHI.

\subsection{Dataset}

Data were collected at the sleep laboratory within a university hospital between September 2025 and September 2026. Each overnight recording included full polysomnography (PSG), thermal video captured by ApneaEye, and clinical annotations of sleep and respiratory events.

\subsubsection{\textbf{Ethical Considerations}}

This study was approved by the Institutional Review Board (IRB) of the Anonymous Medical School and the Institutional Ethics Committee of the Anonymous Institute of Technology. All participants provided written informed consent before participation. Thermal recordings and physiological signals were collected in accordance with institutional guidelines for research involving human participants. All data were anonymized before analysis, and access was restricted to authorized members of the research team. 

\subsubsection{\textbf{Hardware Details}}

\paragraph{\textbf{Polysomnography}:}
Overnight PSG recordings were recorded using the SOMNO HD system (SOMNOmedics) and scored using DOMINO software. The PSG system records multiple physiological signals, including airflow, respiratory effort, blood oxygen saturation, electroencephalography (EEG), and electrocardiography (ECG), enabling comprehensive sleep assessment. Recordings were exported in European Data Format (EDF) or plain-text format for subsequent analysis. For this study, we used the nasal thermistor (32~Hz), nasal pressure signal recorded using a pressure cannula (256~Hz), thoracic and abdominal respiratory inductance plethysmography (RIP) belts and their sum (32~Hz), finger pulse oximetry (SpO$_2$, 4~Hz), and body-position channels. These signals supported respiratory event detection, AHI estimation, and form the basis for evaluating ApneaEye against conventional PSG measurements.

\begin{figure}
    \centering
    \includegraphics[width=0.90\linewidth]{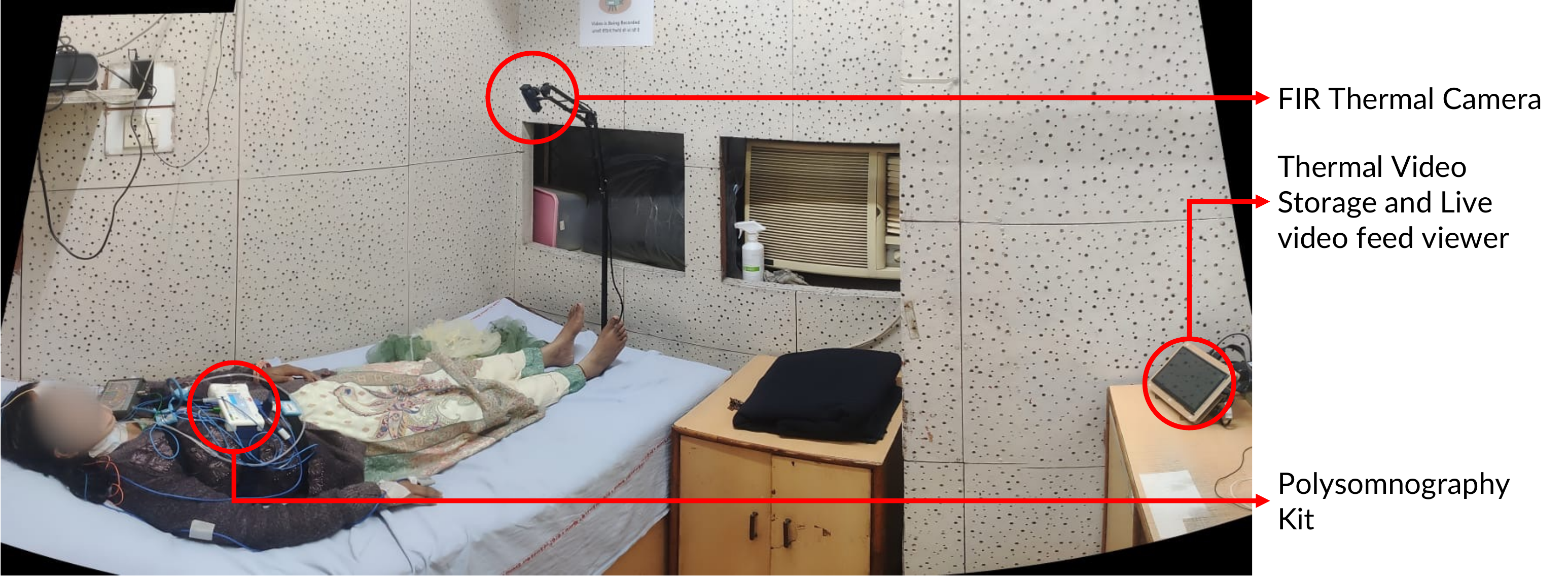}
    \caption{Overnight sleep study setup for ApneaEye evaluation alongside polysomnography (PSG). The participant is instrumented with PSG sensors for clinical reference measurements. The red annotations highlight the thermal camera mounted at the foot of the bed, the PSG recording equipment, and the thermal camera's placement relative to the participant.}
    \label{fig:ApneaEyedeployment} 
\end{figure}

\paragraph{\textbf{Thermal Camera:}}
ApneaEye uses a commercially available long-wave infrared (LWIR) thermal camera (Topdon TC001\footnote{\url{https://www.topdon.com/products/tc001}}) to monitor the sleeping participant. The camera features a $256 \times 192$ pixel thermal detector with a pixel pitch of $12~\mu$m and a $3.2$~mm focal-length lens, providing a field of view of $56^\circ \times 42^\circ$ (horizontal $\times$ vertical). It operates in the $8$--$14~\mu$m spectral range, with a thermal sensitivity (NETD) of less than $40$~mK, and captures thermal video at a frame rate of $25$~Hz. It is positioned at the foot of the bed to maintain a clear view of both the nostril and thoracoabdominal regions (Figure~\ref{fig:ApneaEyedeployment}). The camera is connected to a Raspberry Pi 5, which records the thermal video stream. To limit data loss in the event of recording interruptions, videos are saved in 10-minute segments. A live preview of the thermal feed is provided to clinicians to allow them to verify that the recording is functioning correctly during the study. The Raspberry Pi is time-synchronized with the PSG system to facilitate temporal alignment between thermal videos and physiological signals. 

\subsubsection{\textbf{Ground-Truth Annotation}}

Apnea and hypopnea events were annotated from PSG recordings by junior sleep physicians following the American Academy of Sleep Medicine (AASM) \cite{berry2012aasm,berry2014aasm,berry2017aasm} scoring guidelines. The annotations were subsequently reviewed by a senior sleep physician. Each respiratory event was assigned start and end timestamps. These annotations served as the ground truth for evaluating respiratory event detection and estimating the Apnea-Hypopnea Index (AHI), defined as the number of apnea and hypopnea events per hour of sleep.

\subsubsection{\textbf{Study Cohort and Exclusion Criteria}}\label{sec:excl}
A total of 316 participants were initially enrolled, with corresponding overnight recordings and thermal videos collected. Of these, 55 participants were excluded due to data quality issues or insufficient recording duration. Specifically, 11 recordings were excluded because of CPAP use, 23 because the face was visible for less than 30\% of the total recording duration, 15 due to problems with thermal recording, and 6 because the total sleep duration was less than 2 hours. Following these exclusions, 261 participants were retained in the final study cohort. The characteristics of the final cohort are summarized in Table~\ref{tab:cohort-statistics}. Across these 261 recordings, ApneaEye captured a total of 2006.8~h of thermal video. These quality checks can be automated to help the user in the home setup.

\subsection{Extracting Respiration}

We first employ a region-of-interest (ROI) localizer to identify the nasal and thoracoabdominal regions in the thermal video. These localized regions are then used to extract a thermal proxy for nasal airflow and track thoracoabdominal motion over time, yielding the two respiratory signals used for subsequent apnea and hypopnea detection.

\subsubsection{\textbf{Region of Interest (ROI) Detection}}

To localize the nasal and thoracoabdominal regions, we employ a two-stage object detection pipeline. The first stage detects the participant's face and thoracoabdominal region, while the second stage detects the nostril region within the detected facial region. This hierarchical design constrains nostril localization to the facial region, reducing the search space for the second-stage detector.

Each thermal frame, originally captured at a resolution of $256 \times 192$ pixels, is resized to $640 \times 640$ pixels before being passed to the first-stage detector. For nostril localization, we extract a square crop centered on the detected facial bounding box, with a side length equal to 1.6 times the larger dimension of the bounding box. The crop is then resized to $640 \times 640$ pixels and passed to the second-stage detector. The additional margin around the detected face makes nostril localization more robust to slight inaccuracies in the first-stage facial detection boundaries, reducing the likelihood of excluding the nostrils from the crop. The resulting nostril detections are mapped back to the coordinates of the original thermal frame, and the highest-confidence detection is retained.

To reduce computational overhead, detection is performed every second. The detected bounding boxes are temporally stabilized using a running median filter centered over 11 successful detector outputs to suppresses short-lived bounding-box fluctuations and outlier detections. The filtered detections propagated to intermediate frames where detection is not performed. If a detection fails, the last successful bounding box is retained for at most 2~s to bridge brief detection failure. Beyond this period, the nostril region is treated as unavailable.

The localization model is trained using manually annotated thermal images collected from 20 participants in an earlier pilot study conducted under the same Institutional Review Board (IRB) approval as the main study. These participants are separate from the 316 participants included in the apnea and hypopnea evaluation cohort. Importantly, the ROI annotations are used exclusively to train and evaluate the localization model and are independent of the respiratory event annotations used in the subsequent analysis. To minimize redundant annotations, we initially sample one frame every 30 seconds and apply frame-difference analysis to remove highly similar images. The remaining frames are manually annotated with bounding boxes around the nostril and thoracoabdominal regions. We evaluate the localization model using four-fold participant-grouped cross-validation.


\subsubsection{\textbf{Nasal Respiration}: }
To estimate nasal airflow, we analyze cyclical temperature variations within the localized nostril region. During inhalation, cooler ambient air entering the nasal cavity lowers the temperature near the nostrils, whereas exhalation produces warmer thermal signatures. These periodic temperature changes captured by the thermal camera reflect the breathing cycle and can serve as a thermal proxy for nasal airflow.

A detected bounding box contains not only the airflow pixels from the nostril region but also the surrounding skin, moving edges, and the background. Simply averaging the entire region hence introduces noise from the signals unrelated to airflow. We therefore derive two nasal signals from a stabilized patch around the nostrils: a learned readout that estimates the amplitude of each breath, and the mean of the few pixels with the most activity in the breathing frequency range.

\paragraph{\textbf{Nasal Patch extraction and stabilization}: }
Given the nasal bounding box obtained from the localizer, we extract a $48 \times 48$-pixel patch from the original frame. The patch is centered at the median of the center of the localized bounding box over a 31-second interval and is moved only when this median center drifts by more than 3 pixels, so that small frame-to-frame variations do not constantly move the patch. The patch is sampled at 5~Hz by averaging 5 consecutive frames and is considered valid if the bounding box position remains within 3 pixels of the patch center in at least four of these five frames.

A stretch is defined as a period over which the patch does not move. We keep stretches that last at least 60~s and are valid for at least half of that duration. Let $x_p(t)$ denote the intensity of pixel $p$ at 5~Hz sample $t$ of a stretch, and $\mathcal{T}$ the valid samples. After linearly interpolating the invalid samples, we band-pass each pixel's fluctuation around its mean,
\begin{equation}
\tilde{x}_p = \mathcal{B}\left(x_p - \bar{x}_p\right),
\label{eq:pixel-bp}
\end{equation}
where $\bar{x}_p$ is the pixel's mean intensity over the stretch and $\mathcal{B}$ is a zero-phase third-order Butterworth band-pass filter (0.1--0.7~Hz). Both nasal signals below build on $\tilde{x}_p$. The interpolated values remain marked as invalid and are not used as model inputs

\paragraph{\textbf{Learned nasal readout}: } 

The learned nasal readout is a small network that estimates, from the nasal patch, how large each breath is relative to the night's typical breath. Its input is the central $32 \times 32$-pixel crop of the patch, as two channels: the band-passed fluctuation $\tilde{x}_p(t)$ of every pixel (Eq.~\ref{eq:pixel-bp}) and the stretch-level mean thermal image, repeated at every timestep to provide spatial context about the nostril region. A small convolutional encoder summarizes each 5~Hz sample, and dilated temporal convolutions combine these summaries over $\pm 6$~s. The training target comes from the PSG nasal pressure: the 5-s root-mean-square envelope $e(t)$ of the square-root-transformed, band-passed pressure signal, divided by its median over the night's valid samples.


 The network is trained to minimize the squared error of its prediction on a logarithmic scale. Its output is therefore a 5~Hz breath amplitude rather than an airflow waveform. The readout is trained with patient-grouped three-fold cross-validation, and each night's readout comes only from models that never saw any night of the same patient during training.


\paragraph{\textbf{Breathing-sensitive pixel extraction}: } 

An average bounding box from the localizer is $9 \times 6$~px, of which only a subset (approximately 12 pixels) belong to the nostrils, and the rest are edges and skin that move slightly with respiratory effort, so averaging the whole box also records this movement noise. We therefore keep the six pixels with the strongest breathing-band fluctuation.

Within each stretch, we consider the localizer bounding box enlarged by 4~px on all sides, denoted $\mathcal{R}_n^{+}$, and score each pixel by the root-mean-square (RMS) value of its band-passed fluctuation (Eq.~\ref{eq:pixel-bp}) over the valid samples. The six pixels of $\mathcal{R}_n^{+}$ with the highest RMS form the set $\mathcal{S}_6$, and their mean original intensity estimates the respiratory signal. Where the top-6 signal cannot be extracted because a stretch does not meet the above criteria, the average intensity of the localized nasal bounding box is substituted. 






\subsubsection{\textbf{Thoracoabdominal Movements} :} Thoracoabdominal movements are estimated by tracking motion in the localized chest region. Let $\mathcal{R}_c$ denote the set of pixel coordinates corresponding to the chest region identified by the ROI localization module. We apply Gunnar Farnebäck dense optical flow to compute the motion field between consecutive frames within this region. Let $\mathbf{F}_t(x,y)$ denote the optical flow vector at pixel $(x,y)$ between frames $t-1$ and $t$.

The thoracoabdominal movement signal is computed as the mean magnitude of the optical flow vectors within the chest region:

\begin{equation}
s_t^{c} = \frac{1}{|\mathcal{R}_c|} \sum_{(x,y)\in \mathcal{R}_c} \|\mathbf{F}_t(x,y)\|,
\end{equation}

where $|\mathcal{R}_c|$ denotes the number of pixels in the chest region. The sequence $\{s_t^{c}\}$ forms a time series representing a thermal proxy for the thoracoabdominal respiratory effort.

\subsubsection{\textbf{Refining Respiration} :}  To refine the respiration signals, we apply a band-pass filter corresponding to breathing rates between 6 and 42 breaths per minute (0.1 to 0.7 Hz). 
Additionally, binary validity channels are created for the corresponding signal proxies, with 1 indicating a valid sample and 0 indicating a missing or invalid sample. Invalid samples are passed as zeros in the proxies, and the validity channels are passed to the segmentation model with their corresponding signals.



\subsection{Event Segmentation}
ApneaEye formulates respiratory event detection as a per-second segmentation task, enabling precise temporal localization of individual events rather than assigning a single label to each fixed-length window. As previously discussed in section~\ref{sec:relatedworks}, window-based apnea estimation has limitations~\cite{KristiPortablePSG}. Consequently, AHI estimation depends not only on classification performance but also on the choice of window duration. These limitations motivate our shift from window-based classification to temporal segmentation. By leveraging the extracted nasal airflow and thoracoabdominal motion proxies, ApneaEye identifies respiratory events at finer temporal resolution, enabling both event-level localization and more direct AHI estimation through event counting.

A one-dimensional U-Net processes long windows of extracted respiratory signals, optionally including oxygen saturation (SpO$_2$), and predicts the probability of respiratory events every second. The resultant probability is converted to discrete events to allow per-event statistics as well as AHI estimation.




\subsubsection{\textbf{Model Description}}
A one-dimensional U-Net with 2.3 million trainable parameters segments respiratory events at one-second. The 10~Hz respiratory inputs are passed through successive strided convolutional layers to reduce the temporal resolution to 1Hz. A U-Net encoder-decoder then captures temporal context through progressive downsampling and upsampling, with skip connections preserving information needed for event localization. Each convolutional block comprises a one-dimensional convolution, Group Normalization, GELU activation, and dropout. 

The model receives three camera-derived respiratory signals: the breathing-sensitive pixels' nasal signal, the learned nasal breath-amplitude readout, and the thoracoabdominal motion signal. Binary validity channels accompany the corresponding signals to indicate sample availability. When utilized, SpO$_2$ is processed by a separate encoder and then merged with the reduced temporal resolution respiratory signals at 1Hz, and then passed to the U-Net. The output head produces two logits per second for background and respiratory-event classes

\subsubsection{\textbf{Training Procedure}}
The model is trained on 600~s windows with stride of 175~s. We use three-fold cross validation, with holding out one of the folds, and splitting the rest into 80:20 Training:Validation split. To improve robustness to the signal variability and polarity inversions, each respiratory input is subjected to random sign inversions, gain scaling (uniformly sampled from 0.8 to 1.25), an zeroing. Validity and SpO$_2$ channels are not augmented. 

The training objective is class-weighted per-second cross-entropy, with weights inversely proportional to the square root of class prevalence, normalized so that the smaller weight is one and capped at 12. The model is trained using AdamW optimizer, with learning rate of 0.0003, a one-cycle schedule, weight decay of 0.0001, gradients clipped at 5, batch size of 16, and at half-precision (bf16), for upto 60 maximum passes of training set, with early stopping after 12 epochs without improvement in validation area under the precision--recall curve (AUPRC). The checkpoint with the best validation AUPRC is retained.

\subsubsection{\textbf{Inference and Postprocessing}}
At inference, 600~s windows with 300~s overlaps are evaluated, with the predictions from overlapping windows averaged to obtain a single probability vector at 1~Hz resolution. The validation nights are used to select the probability threshold for each run, as optimal thresholds vary between training runs even on the same fold.

The binary time series of predicted events is converted to event intervals by merging detections separated by 1~s or less. Events smaller than 10~s are discarded and exceeding 120~s are split at low-probability points, subject to minimum duration constraints. 

\subsubsection{\textbf{Evaluation and Scoring}}
The model is evaluated on all 261 participants via cross-validation. 
For event-level evaluation, predicted events are processed chronologically and matched to the unmatched reference event with the highest IoU, provided that IoU is at least 0.3. The AHI is calculated as total number of events divided by total sleep time in hours. We also report per-second weighted F1 and macro F1, computed over sleep seconds

\subsection{Event Typing}

To classify respiratory events, we process a 120-s window centered on each event. A encoder with the same architecture as the front part of the segmentation pipeline downsamples the 10Hz breathing channels to 1Hz and optionally adds the 1Hz SpO2 channel, producing a 128-dimensional representation per second for 120 seconds. The architecture involves four temporal blocks, followed by dropout, an attention mechanism, and a classification head with a 128-unit fully connected layer, and a final two-unit layer to predict apnea and hypopnea scores. During training, the windows are augmented with shifts of $\pm 10$~s, polarity inversions, and random gain.

\begin{table}[t]
\centering
\caption{Summary of the study cohort and recording statistics (261 nights). Respiratory events were scored by sleep physicians, with event midpoints occurring during sleep. Hypopneas required a $\geq$3\% oxygen desaturation.}
\label{tab:cohort-statistics}

\small
\renewcommand{\arraystretch}{1.1}
\setlength{\tabcolsep}{5pt}

\begin{tabularx}{\linewidth}{@{}X r X r@{}}
\toprule

\multicolumn{4}{@{}l}{\textbf{Study Cohort}} \\
\addlinespace[2pt]
Nights (recordings) & 261
& Gender (male / female) & 138 / 123 \\

Age, years (mean, range) & 45.19 (18--76)
& Weight, kg (mean, range) & 74.91 (46--104.7) \\

Nights with sleep apnea (AHI $\geq$ 5) & 230
& Height, cm (mean, range) & 167.25 (143--188) \\

Mean AHI / Median AHI (range) & 29.4 / 22.0 (0.3--113.0)
& Severity (normal / mild / moderate / severe) & 31 / 69 / 63 / 98 \\

\midrule

\multicolumn{4}{@{}l}{\textbf{Recording Statistics}} \\
\addlinespace[2pt]
Total thermal video & 2006.8 h
& Mean duration per night (range) & 7.69 h (3.83--9.79) \\

\midrule

\multicolumn{4}{@{}l}{\textbf{Respiratory Events}} \\
\addlinespace[2pt]
Obstructive apnea & 16,465
& Central apnea & 525 \\

Mixed apnea & 540
& Hypopnea & 27,726 \\

\addlinespace[2pt]
\textbf{Total respiratory events} & \textbf{45,256}
& & \\

\bottomrule
\end{tabularx}
\end{table}

\subsection{Sleep Time Estimation}

Sleep and wake states were estimated from a body-movement activity signal derived from the bedside thermal video. Each recording ($256 \times 192$ pixels, 25 frames/s) was sampled at 5 Hz. Movement was quantified as the mean absolute pixel-wise difference between consecutive standardised frames (separated by 0.2 s), averaged across all pixels, yielding a video-based analogue of actigraphy activity counts. The resulting signal was normalised within each night and segmented into 30-s epochs, aligned with the corresponding annotation obtained from gold standard PSG. Each epoch was assigned a binary wake (W) or sleep (N1, N2, N3, or REM) label based on PSG annotations. We employed a sequence model comprising a per-epoch one-dimensional convolutional encoder followed by a bidirectional gated recurrent unit (BiGRU) network to capture temporal dependencies across the night. 
At inference, the entire night was processed in a single forward pass. Models were trained using class-weighted cross-entropy, with class weights proportional to the square root of the inverse class frequencies. 
A proportion of the training nights was reserved for early stopping based on Cohen's $\kappa$, evaluated every 250 steps with a patience of 12 evaluations and the checkpoint achieving the highest $\kappa$ was retained. Performance was evaluated using five-fold cross-validation grouped by patient across 261 participants, ensuring that each night was evaluated by a model that had not been trained on data from the corresponding patient. Each configuration was trained using three random seeds. Total sleep time (TST) was calculated by summing the predicted and PSG-labelled sleep epochs.

\subsection{Body Pose Estimation}\label{sec:pose}

We trained image classifiers to estimate body position from individual thermal frames, using three classes: supine, left and right. One frame was sampled every 30~s from each recording and labelled using the synchronised PSG body-position channel, yielding 272{,}699 labelled frames. We compared eight architectures fine-tuned end to end: ResNet-18 and ResNet-50~\cite{resnet}, EfficientNet-B0~\cite{efficientnet}, MobileNetV3-Large~\cite{mobilenetv3}, ConvNeXt-Tiny~\cite{convnext}, YOLOv8n and YOLO11n classification networks~\cite{yolov8}, and DINOv2 ViT-S/14~\cite{dinov2}, replacing their classification heads with three-class output layers. Each frame was resized to the model's required input resolution. During training, we applied random zoom (0.85--1.0), translation ($\pm8\%$), rotation ($\pm10^\circ$), contrast adjustment (0.8--1.2), and horizontal flipping with left-right label swapping. Models were trained using class-weighted cross-entropy with weights proportional to the inverse square root of class frequency, and label smoothing of 0.05. 
To account for temporal continuity in body position, we smoothed the frame-level probabilities using a hidden Markov model (HMM). All models were evaluated using patient-grouped three-fold cross-validation, consistent with the event segmentation model. Predictions were pooled across test folds and evaluated on PSG-scored sleep, comprising 215{,}317 frames from 261 nights. We report overall accuracy and class-wise precision, recall, and F1 scores for supine, left, and right positions ).

\section{Results and Evaluations}
Table~\ref{tab:cohort-statistics} summarizes the demographic characteristics, clinical composition, recording coverage, and respiratory event distribution of the evaluation cohort.
All predictions are evaluated against respiratory events annotated by sleep physicians from PSG signals following AASM guidelines, which serve as the clinical reference throughout.
To establish a performance ceiling, we also pass PSG contact sensor signals through the same event segmentation model — this PSG contact sensor configuration isolates the contribution of signal quality from that of the model. Notably, both configurations are evaluated against the same human annotations, which carry their own variability: physicians annotating two six-to-ten hour studies per day face a cognitively demanding task where borderline events introduce irreducible disagreement. 
The gap between ApneaEye and the PSG contact sensor baseline therefore reflects both signal quality differences and annotation noise in the reference itself.



\subsection{Apnea Detection}
\begin{table}
\centering
\caption{Comparison of ApneaEye thermal sensing and PSG contact sensors across 261 patients for respiratory event detection, AHI estimation, and sleep apnea severity classification. Metrics are reported at the event, AHI, and severity levels. Reported metrics are Macro-F1 (MF1), Weighted-F1 (WF1), Mean Absolute Error (MAE), Intra-class Correlation Coefficient (ICC), Accuracy of classifying in the same and within-one severity band. $\uparrow$ indicates that higher values are better, while $\downarrow$ indicates that lower values are better.}
\label{tab:apneaeye-psg-comparison}
\setlength{\tabcolsep}{4pt}
\resizebox{\textwidth}{!}{%
\begin{tabular}{@{}l*{12}{c}@{}}
\toprule
& \multicolumn{6}{c}{\textbf{ApneaEye (Thermal Camera)}}
& \multicolumn{6}{c}{\textbf{PSG Contact Sensors}} \\
\cmidrule(lr){2-7} \cmidrule(lr){8-13}

& \multicolumn{2}{c}{\textbf{Event-Level}}
& \multicolumn{2}{c}{\textbf{AHI-Level}}
& \multicolumn{2}{c}{\textbf{Severity-Level}}
& \multicolumn{2}{c}{\textbf{Event-Level}}
& \multicolumn{2}{c}{\textbf{AHI-Level}}
& \multicolumn{2}{c}{\textbf{Severity-Level}} \\
\cmidrule(lr){2-3} \cmidrule(lr){4-5}
\cmidrule(lr){6-7} \cmidrule(lr){8-9}
\cmidrule(lr){10-11} \cmidrule(lr){12-13}

\textbf{Input}
& \textbf{MF1$\uparrow$} & \textbf{WF1$\uparrow$}
& \textbf{MAE}$\downarrow$ & \textbf{ICC$\uparrow$}
& \makecell[c]{\textbf{Exact}\\ \textbf{Acc}$\uparrow$} & \makecell[c]{ \textbf{$\pm$1 Band}\\ \textbf{Acc}$\uparrow$}
& \textbf{MF1$\uparrow$} & \textbf{WF1$\uparrow$}
& \textbf{MAE}$\downarrow$ & \textbf{ICC$\uparrow$}
& \makecell[c]{\textbf{Exact}\\ \textbf{Acc}$\uparrow$} & \makecell[c]{ \textbf{$\pm$1 Band}\\ \textbf{Acc}$\uparrow$} \\
\midrule

Thorac Motion
& 0.739 & 0.801 & 11.5 & 0.780 & 0.534 & 0.893
& 0.766 & 0.824 & 9.6 & 0.862 & 0.611 & 0.931 \\

Nasal Airflow
& 0.704 & 0.769 & 11.4 & 0.780 & 0.600 & 0.897
& 0.814 & 0.863 & 8.4 & 0.888 & 0.648 & 0.944 \\


ApneaEye (Nasal + Thorac)
& 0.751 & 0.809 & 9.7 & 0.858 & 0.621 & 0.933
& 0.814 & 0.863 & 8.4 & 0.887 & 0.649 & 0.943 \\

Nasal + Thorac + PSG SpO$_2$
& 0.804 & 0.848 & 6.8 & 0.919 & 0.699 & 0.979
& 0.834 & 0.877 & 6.9 & 0.911 & 0.703 & 0.973 \\

\bottomrule
\end{tabular}%
}
\end{table}
ApneaEye use nasal airflow and thorac motion signals to segment out the respiratory events. Respiratory events are then counted to estimate the Apnea-Hypopnea Index (AHI). Table~\ref{tab:apneaeye-psg-comparison} summarizes performance at the event, AHI, and severity levels. For event detection, a predicted event was considered a true positive if its temporal overlap with a ground-truth event exceeded a threshold of 0.3.

Using both nasal airflow and thoracoabdominal motion extracted by ApneaEye, we achieve a macro-F1 (MF1) of 0.751 and a weighted-F1 (WF1) of 0.809 for respiratory event detection. The resulting Apnea-Hypopnea Index (AHI) estimates have a mean absolute error (MAE) of 9.7 events/hour and an intraclass correlation coefficient (ICC) of 0.858. For four-class sleep apnea severity classification, ApneaEye achieves an exact classification accuracy of 62.1\%, with 93.3\% of nights classified within one severity band of the reference. Using thoracoabdominal motion alone yields an AHI MAE of 11.5 events/hour, whereas incorporating nasal airflow reduces the MAE to 9.7 events/hour. This improvement highlights the value of combining complementary respiratory signals for estimating sleep apnea severity.

\begin{figure}
    \centering
    \includegraphics[width=1\linewidth]{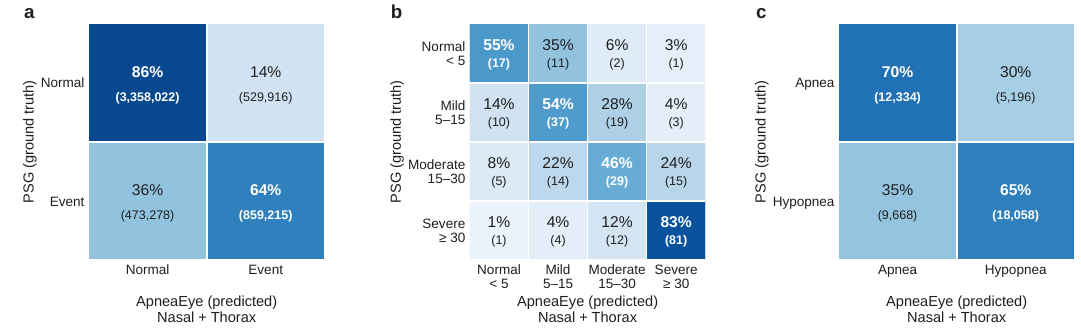}
    \caption{Comparison of ApneaEye predictions against PSG annotations. (a) Confusion matrix for normal and respiratory event detection. (b) Distribution of severity predictions across PSG-defined AHI severity categories (events/hour). (c) Confusion matrix for apnea and hypopnea classification. Percentages are computed row-wise, with absolute counts shown in parentheses.}
    \label{fig:confusion_matrices}
\end{figure}

\begin{figure}
    \centering
    \includegraphics[width=0.9\linewidth]{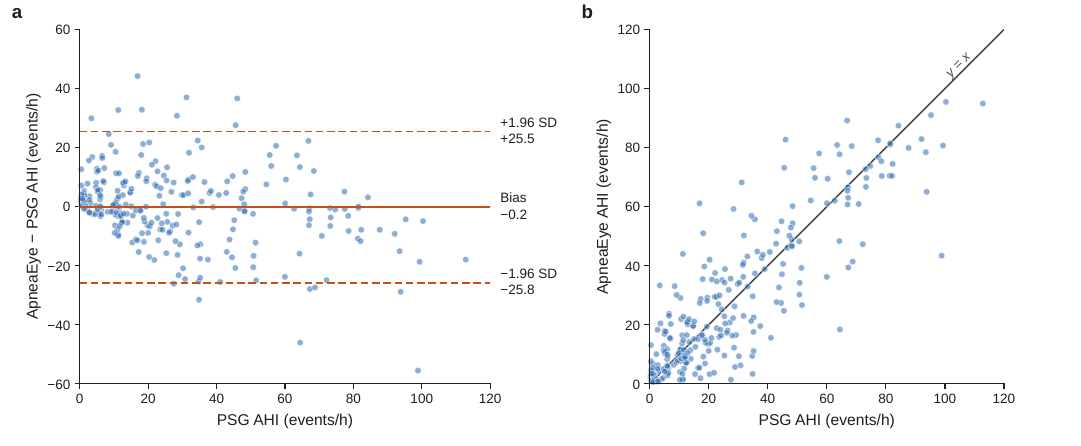}
    \caption{Agreement between ApneaEye-derived and polysomnography (PSG)-measured apnea--hypopnea index (AHI). ApneaEye extracts nasal airflow and thoracoabdominal motion from thermal imagery to estimate AHI. (a) Modified Bland--Altman plot showing the difference between ApneaEye and PSG AHI as a function of PSG AHI, with a mean bias of $-0.2$ events/h and 95\% limits of agreement from $-25.8$ to $25.5$ events/h. (b) Scatter plot comparing ApneaEye and PSG AHI; the solid diagonal line represents the line of identity ($y=x$).}
    \label{fig:AHI_Agreement_Cam}
\end{figure}

We further compare ApneaEye against models using respiratory signals acquired by PSG contact sensors. These comparisons provide a reference for the performance achievable when respiratory signals are measured directly using conventional contact-based sensors, rather than indirectly reconstructed from thermal imagery. Using thoracoabdominal motion alone, the PSG-based configuration achieves an AHI MAE of 9.6 events/hour, compared with 11.5 events/hour for ApneaEye. When nasal airflow is incorporated, the MAE decreases to 8.4 events/hour for PSG and 9.7 events/hour for ApneaEye. \textit{Thus, despite relying on non-contact thermal sensing to reconstruct respiratory signals, ApneaEye approaches the performance of the corresponding PSG-based configuration.} This comparison helps quantify the performance gap between contact-based respiratory measurements and their non-contact counterparts.

Finally, we investigate the benefit of incorporating peripheral oxygen saturation (SpO$_2$). Combining ApneaEye-derived nasal airflow and thoracoabdominal motion with SpO$_2$ reduces the AHI MAE to 6.8 events/hour and increases the ICC to 0.919. The resulting exact severity classification accuracy is 69.9\%, with 97.9\% of nights classified within one severity band of the reference. These results are comparable to those obtained using PSG respiratory signals together with SpO$_2$, which yield an AHI MAE of 6.9 events/hour and an ICC of 0.911. At the event level, this multimodal configuration also achieves an MF1 of 0.804 and a WF1 of 0.848 using ApneaEye-derived respiratory signals. \textit{Overall, these findings demonstrate that combining nasal airflow and thoracoabdominal motion improves AHI estimation over thoracoabdominal motion alone, while incorporating SpO$_2$ brings the performance of the ApneaEye-based configuration closer to that of the PSG-based configuration across event detection, AHI estimation, and severity classification.}

The modified Bland-Altman analysis (Figure~\ref{fig:AHI_Agreement_Cam}a) shows a mean bias of $-0.2$ events/hr, indicating no systematic over- or under-estimation, within $95\%$ limits of agreement of $-25.8$ to $+25.5$ events/hr. These limits of agreement indicate uncertainty in individual AHI estimates, particularly for cases near clinical severity thresholds. We therefore additionally evaluate monitoring performance at clinically meaningful AHI cutoffs.
Importantly, at these severity levels, large absolute errors have limited clinical consequence, as whether a patient's AHI is 60 or 80, the diagnosis and treatment path are the same. For the clinical task of severity staging, the ±1 band accuracy of 93.3\% is therefore the more informative metric as it shows that ApneaEye places nearly all patients within one category of the correct severity class, which is the decision that determines clinical management.

Figure~\ref{fig:confusion_matrices} summarizes ApneaEye's predictions across three classification tasks. For binary event classification (Figure~\ref{fig:confusion_matrices}a), ApneaEye correctly identifies 64\% of respiratory events while missing 36\%. This seemingly large false negative rate for events does not translate into a reduced sensitivity for detecting disease or its severity, because AHI is computed from event counts across the full night, where missed short events have limited impact on the total. 
For four-class severity classification (Figure~\ref{fig:confusion_matrices}b), errors are concentrated at adjacent severity boundaries.
For classifying event type (Figure~\ref{fig:confusion_matrices}), ApneaEye correctly classifies 70\% of apnea events and 65\% of hypopnea events. The lower hypopnea accuracy is expected: the AASM definition of hypopnea requires a confirmed oxygen desaturation, which thermal imaging cannot directly measure. Without SpO$_2$ measurements, the system must infer hypopnea from airflow reduction alone, making ambiguous borderline events harder to distinguish from normal breathing. This explains why SpO$_2$ augmentation improves event-level performance.


\begin{table}[t]
\centering
\caption{AHI estimation by nostril visibility. Nights are grouped by the share of sleep during which the nostrils were visible to the camera (Table~\ref{tab:nostril-visibility}; the single night below 30\% is merged into the lowest group). N: nasal airflow; T: thoracoabdominal motion. MAE is in events/h; exact severity is the share of nights assigned the correct AASM severity class.}
\label{tab:ahi-by-visibility}
\begin{tabular}{l r ccc ccc}
\toprule
 & & \multicolumn{3}{c}{\textbf{AHI MAE (events/h) $\downarrow$}} & \multicolumn{3}{c}{\textbf{Exact severity (\%) $\uparrow$}} \\
\cmidrule(lr){3-5}\cmidrule(lr){6-8}
\textbf{\makecell[l]{Nostrils visible\\(share of sleep)}} & \textbf{Nights}
& \textbf{\makecell{Camera\\T}}
& \textbf{\makecell{Camera\\N+T}}
& \textbf{\makecell{Camera\\N+T+SpO$_2$}}
& \textbf{\makecell{Camera\\T}}
& \textbf{\makecell{Camera\\N+T}}
& \textbf{\makecell{Camera\\N+T+SpO$_2$}} \\
\midrule
$<$50\%      & 46  & 10.10 & 9.22  & 7.42 & 51.1 & 53.3 & 63.0 \\
50--70\%     & 48  & 15.75 & 12.08 & 7.40 & 46.9 & 57.3 & 71.9 \\
70--90\%     & 80  & 10.52 & 8.95  & 5.56 & 56.9 & 70.6 & 78.1 \\
$\geq$90\%   & 87  & 10.85 & 9.45  & 7.18 & 55.2 & 61.5 & 64.9 \\
\midrule
\textbf{All} & \textbf{261} & 11.52 & 9.74  & 6.77 & 53.4 & 62.1 & 69.9 \\
\bottomrule
\end{tabular}
\end{table}

\subsection{AHI estimation for varying nostril visibility}

We further investigate whether the benefit of nasal airflow depends on how consistently the nostrils are visible throughout the night. Table~\ref{tab:ahi-by-visibility} reports AHI estimation performance across four groups of nights, stratified by the fraction of sleep during which the nostrils were visible. Incorporating nasal airflow improves AHI estimation across all four groups compared with using thoracoabdominal motion alone. The AHI MAE decreases from 10.10 to 9.22 events/h in nights with less than 50\% nostril visibility, from 15.75 to 12.08 events/h for 50--70\% visibility, from 10.52 to 8.95 events/h for 70--90\% visibility, and from 10.85 to 9.45 events/h for visibility above 90\%. A similar improvement is observed in exact severity classification, with gains across all visibility groups. Notably, even when the nostrils are visible during less than half of sleep, incorporating nasal airflow improves both AHI estimation and severity classification.

This benefit despite limited overall visibility may partly be explained by the distribution of respiratory events across body positions. As shown in Table~\ref{tab:nostril-visibility}, 74.7\% of scored respiratory events occur during supine sleep, which accounts for 66.6\% of total sleep time. Importantly, nostril visibility is higher during supine sleep (84.2\%) than during left (62.6\%) or right lateral sleep (54.2\%). Thus, although nasal airflow is not continuously available throughout the night, it would be available during a substantial proportion of respiratory events. \textit{Overall, these results suggest that incorporating nasal airflow provides complementary information for AHI estimation even when nostril visibility is intermittent, rather than requiring continuous visibility throughout the night.}

\begin{table}[t]
\centering
\caption{Respiration rate (measured in Breaths per minute or BrPM) estimation accuracy of ApneaEye against PSG reference sensors. $\downarrow$ indicates that lower is better.}
\label{tab:respiration-rate}
\begin{tabular}{llccc}
\toprule
\textbf{ApneaEye Signal} &
\textbf{PSG Reference Sensor} &
\makecell[c]{\textbf{Median of} \\ \textbf{Absolute Errors} \\ \textbf{(BrPM) $\downarrow$}} &
\makecell[c]{\textbf{Mean of} \\ \textbf{Absolute Errors} \\ \textbf{(BrPM) $\downarrow$}} &
\makecell[c]{\textbf{Root Mean} \\ \textbf{Squared Error} \\ \textbf{(BrPM) $\downarrow$}} \\
\midrule
Nasal Airflow & Oronasal Thermistor & 0.15 & 1.96 & 4.27 \\
Thoracic Motion & Thoracic Movement Sensor & 0.29 & 2.19 & 4.42 \\
\bottomrule
\end{tabular}
\end{table}

\subsection{Respiration Extraction}

We estimated respiration rate (RR) as the dominant breathing frequency in 30-s windows with a 15-s hop during PSG-scored sleep free of respiratory events, and evaluated agreement between ApneaEye-derived signals and their corresponding PSG reference channels across 261 test nights (Table~\ref{tab:respiration-rate}). ApneaEye's extracted nasal airflow was compared against the PSG oronasal thermistor, while its thoracic-motion signal was compared against the PSG thoracic respiratory-effort channels. Across the respective reference channels, the median absolute error was 0.15 breaths per minute (BrPM) for nasal airflow and 0.29 BrPM for thoracic motion. The corresponding mean absolute errors (MAEs) were 1.89--1.96 BrPM for nasal airflow and 2.15--2.19 BrPM for thoracic motion. These MAEs are close to the illustrative target of less than 2 BrPM discussed by Charlton et al.~\cite{Charlton2018_BR_numbers}; however, standardized performance thresholds for RR estimation have not been established. A tolerance of $\pm 2$ BrPM has also been used as an operational definition of measurement accuracy in clinical practice~\cite{keshvani2019_BR_number}. Using this tolerance, 80\% of nasal-airflow estimates and 78\% of thoracic-motion estimates were within $\pm 2$ BrPM of their respective PSG references.

Recent work in ubiquitous computing has investigated RR estimation using unobtrusive sensing modalities. PillowSense estimates RR using computational fabrics integrated into a pillowcase and reports MAEs of 2.4 BrPM in a laboratory study and 3.1 BrPM in a longitudinal study spanning 151 nights~\cite{Shao2025PillowSense}. Kazemi et al.~\cite{Kazemi2025SmartwatchRR} estimated RR from smartwatch photoplethysmography (PPG) and accelerometer data, reporting MAEs of 2.29 and 3.09 BrPM on the PPG-DaLiA and WESAD datasets. Liu et al.~\cite{liu2023contactless} developed a video-based, contactless RR monitoring approach for ICU patients and reported an MAE of 2.81 BrPM and an RMSE of 3.63 BrPM in one evaluation configuration. Although these studies provide useful context, differences in sensing modalities, populations, and evaluation protocols preclude direct ranking based on error alone. \textit{ApneaEye achieves competitive RR estimation errors while being evaluated against PSG reference channels across 261 test nights.}


\subsection{ROI Localizer}

\begin{table}[t]
  \centering
  \caption{Localization performance of ROI detector model for nostril and thorax regions. Since nostrils are extremely small objects (average bounding-box size of $9 \times 6$ pixels), their performance is evaluated at IoU@0.25, whereas thorax localization is evaluated at the standard IoU@0.5 threshold. $\uparrow$ indicates that higher values are better.}
  \label{tab:s-loc-pr}
  \begin{tabular}{@{}lrrrrrr@{}}
    \toprule
    & \multicolumn{3}{c}{\textbf{Nostril (IoU 0.25)}} & \multicolumn{3}{c}{\textbf{Thorax (IoU 0.5)}} \\
    \cmidrule(lr){2-4}\cmidrule(l){5-7}
    \textbf{Model} & \textbf{Precision}$\uparrow$ & \textbf{Recall}$\uparrow$ & \textbf{mAP}$\uparrow$  & \textbf{Precision}$\uparrow$ & \textbf{Recall}$\uparrow$ & \textbf{mAP}$\uparrow$ \\
    \midrule
    YOLOv8n & 0.844 & 0.644 & 0.730 & 0.884 & 0.840 & 0.897 \\   
    YOLOv8m & 0.884 & 0.696 & 0.760 & 0.860 & 0.832 & 0.892 \\   
    YOLO11m & 0.788 & 0.706 & 0.732 & 0.869 & 0.865 & 0.910 \\   
    YOLO12m & 0.863 & 0.758 & 0.802 & 0.881 & 0.860 & 0.916 \\   
    RT-DETR-L & 0.873 & 0.803 & 0.813 & 0.898 & 0.879 & 0.909 \\   
    Faster R-CNN & 0.904 & 0.756 & 0.801 & 0.917 & 0.855 & 0.913 \\   
    \bottomrule
  \end{tabular}
\end{table}

The localization performance of different object detection models is summarized in Table~\ref{tab:s-loc-pr}. For nostril localization, we evaluate performance at IoU@0.25 due to the extremely small target size (average bounding-box size of $9 \times 6$ pixels). As discussed in Appendix~\ref{sec:tiny_object_problem}, even a one-pixel diagonal displacement reduces IoU to 0.59, making conventional IoU@0.5 evaluation overly sensitive to minor localization errors. At IoU@0.25, RT-DETR-L achieves the highest mAP (0.813) and recall (0.803), while Faster R-CNN achieves the highest precision (0.904). For thorax localization, evaluated at the standard IoU@0.5 threshold, YOLO12m achieves the highest mAP (0.916), followed by Faster R-CNN (0.913) and RT-DETR-L (0.909). We selected RT-DETR-L for the subsequent pipeline, prioritizing nostril localization performance given the smaller target size and greater localization difficulty. It achieves the best nostril mAP and recall while maintaining competitive thorax localization performance.

\subsubsection{\textbf{Nostril Visibility}}


\begin{table*}[t]
\centering
\caption{Sleep time, scored respiratory events, and camera visibility by body position (PSG position sensor) across the 261 nights (left), and distribution of nights by the fraction of sleep with visible nostrils (right). Visibility is the fraction of sleep time during which the nostrils or chest were visible to the camera; respiratory events are assigned the body position at their midpoint.}
\label{tab:nostril-visibility}

\begin{minipage}[c]{0.70\textwidth}
\centering
\begin{adjustbox}{max width=\linewidth}
\begin{tabular}{lrrrr}
\toprule
\textbf{Position} & \makecell{\textbf{Sleep (\%)}} & \makecell{\textbf{Events (\%)} }
& \makecell{\textbf{Nostrils}\\\textbf{visible (\%)}} 
& \makecell{\textbf{Chest}\\\textbf{visible (\%)}} \\
\midrule
Supine            & 66.6 & 74.7 & 84.2 & 97.8 \\
Left lateral      & 13.0 & 11.4 & 62.6 & 96.8 \\
Right lateral     & 19.1 & 13.3 & 54.2 & 97.1 \\
Prone             &  0.8 &  0.3 & 38.7 & 100.0 \\
Upright / unknown &  0.5 &  0.3 & 71.2 & 98.2 \\
\midrule
All               & 100  & 100  & 75.3 & 97.6 \\
\bottomrule
\end{tabular}
\end{adjustbox}
\end{minipage}
\hfill
\begin{minipage}[c]{0.25\textwidth}
\centering
\begin{tabular}{lr}
\toprule
\textbf{Nostrils visible} & \textbf{Nights} \\
\midrule
$\geq$90\% of sleep & 87 \\
70--90\%            & 80 \\
50--70\%            & 48 \\
30--50\%            & 46 \\
\bottomrule
\end{tabular}
\end{minipage}
\end{table*}

Nasal airflow can only be measured while the nostrils are in view of the camera. We consider a second of recording nostril-visible when the stabilized nasal patch is valid for at least half of that second. Table~\ref{tab:nostril-visibility} summarizes how long the nostril and thoracoabdominal regions were visible to the camera. Across the 261 nights, the nostrils were visible during 75.3\% of the total sleep time. Per night, the nostrils were visible during a median of 80.9\% of sleep (interquartile range 58.3--94.3\%), corresponding to a median of 4.6~h per night. Visibility exceeded 90\% of sleep in 87 nights (33\%) and 70\% in 167 nights (64\%), while 46 nights (18\%) had visibility below 50\%. Visibility was lower during wakefulness in bed (61.7\%), when participants move and reposition more often. In contrast, the thoracoabdominal region was visible during 97.6\% of sleep, with a median of 100\% per night.

The main cause of lost nostril visibility is that the face is not fully in view rather than a failure of the detector. In a manual review of 119 sampled sleep frames without a nostril detection, drawn from 24 nights, the face was not visible in 42\% of frames (e.g., covered by a blanket, turned away from the camera, or outside the field of view), partly visible in 38\%, and fully visible in 20\%. To handle these gaps, ApneaEye passes a per-second validity channel for each respiratory signal to the segmentation model, which can therefore rely on thoracoabdominal motion when the nostrils are not visible.


\begin{table}
\centering
\caption{Effect of sleep duration estimation on AHI estimation
across respiratory signal configurations ($N=261$). PSG TST
denotes sleep duration derived from polysomnography, while
ApneaEye TST denotes sleep duration estimated using video-based
actigraphy. For both TST estimates, AHI is calculated using
respiratory events detected by ApneaEye from the specified
respiratory channels, after excluding events occurring during
wakefulness based on the corresponding sleep-wake annotations.
Reference AHI, used as the reference for metric computation, is
calculated using physician-annotated respiratory events and
PSG-derived TST. $\uparrow$ indicates higher is better and
$\downarrow$ indicates lower is better.}
\label{tab:tst_impact_ahi}
\small
\setlength{\tabcolsep}{4pt}
\begin{tabular}{lcc cc cc}
\toprule
& \multicolumn{2}{c}{\textbf{AHI MAE} $\downarrow$}
& \multicolumn{2}{c}{\textbf{ICC} $\uparrow$}
& \multicolumn{2}{c}{\textbf{Exact Match} (\%)$\uparrow$} \\
\cmidrule(lr){2-3}
\cmidrule(lr){4-5}
\cmidrule(lr){6-7}
\textbf{Respiratory Signals}
& \textbf{PSG TST} & \textbf{Video TST}
& \textbf{PSG TST} & \textbf{Video TST}
& \textbf{PSG TST} & \textbf{Video TST} \\
\midrule
Nasal
& 11.56 & 11.32 & 0.78 & 0.79 & 57.1 & 57.9 \\

Thorax
& 11.70 & 11.46 & 0.78 & 0.78 & 53.3 & 53.3 \\

Nasal + Thorax
& 9.83 & 9.78 & 0.85 & 0.85 & 59.0 & 58.8 \\

Nasal + Thorax + SpO$_2$
& 6.86 & 7.09 & 0.92 & 0.91 & 70.9 & 66.7 \\
\bottomrule
\end{tabular}
\end{table}

\begin{figure}
    \centering
    \includegraphics[width=0.9\linewidth]{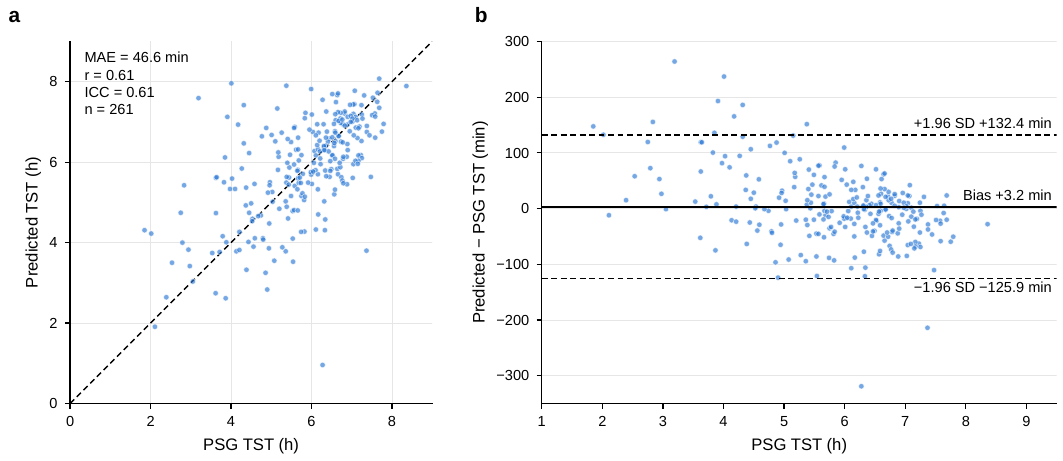}
    \caption{Agreement between video-based and PSG-derived total sleep time (TST) estimates across 261 overnight recordings. (a) Scatter plot of predicted TST against PSG-derived TST, with the dashed line indicating perfect agreement. The model achieves an MAE of 46.6 min, a Pearson correlation coefficient of 0.61, and an intraclass correlation coefficient (ICC) of 0.61. (b) Bland–Altman plot showing the difference between predicted and PSG-derived TST against PSG-derived TST. The mean bias is +3.2 min, with 95\% limits of agreement ranging from -125.9 to +132.4 min.}
    \label{fig:TST_Results}
\end{figure}

\subsection{Total Sleep Time Estimation}

We evaluated the accuracy of video-based total sleep time (TST) estimation against PSG-derived TST across 261 overnight recordings (Fig~\ref{fig:TST_Results}). The proposed method achieves a mean absolute error (MAE) of 46.6 min (95\% CI: 41.2-52.9 min) and a median absolute error of 33.0 min. The estimated TST exhibits a small positive bias of 3.2 min. The Pearson correlation coefficient and intraclass correlation coefficient (ICC(2,1)) are both 0.61, indicating moderate correspondence between video-based and PSG-derived estimates. 

We further investigated the impact of using video-based TST estimates on downstream AHI estimation by replacing PSG-derived TST with ApneaEye-estimated TST while keeping the respiratory event predictions unchanged. As shown in Table~\ref{tab:tst_impact_ahi}, AHI estimation performance remains largely consistent across the two TST sources. For nasal, thoracic, and combined nasal-thoracic configurations, the AHI MAE changes from 11.56 to 11.32, 11.70 to 11.46, and 9.83 to 9.78 events/h, respectively. The corresponding ICC values remain nearly unchanged, as does exact severity classification accuracy. For the SpO$_2$-augmented configuration, the AHI MAE increases from 6.86 to 7.09 events/h, while the ICC remains high (0.92 vs. 0.91) and exact severity classification decreases from 70.9\% to 66.7\%. Overall, these results suggest that replacing PSG-derived TST with video-based estimates has a limited impact on AHI estimation, supporting the feasibility of using video-derived sleep duration in a contactless apnea screening pipeline.

\subsection{Body Position Estimation.}
In this section, we evaluate ApneaEye's ability to estimate body positions during sleep. 
As a camera-based system, ApneaEye can capture body posture without requiring additional wearable sensors. 
This capability is particularly useful for identifying Positional Obstructive Sleep Apnea (POSA), a clinical phenotype in which the severity of obstructive sleep apnea varies with body position.
As shown in Table~\ref{tab:pose}, all models achieve high overall accuracy (0.936--0.947). ConvNeXt-T achieves the highest accuracy of 0.947, closely followed by ResNet18 (0.945). Performance is generally stronger for supine and right positions than for the left position. ConvNeXt-T achieves the highest F1-scores for left (0.897) and right (0.943) positions, while ResNet18 and ConvNeXt-T share the highest supine F1-score (0.964). Overall, the results indicate that thermal images provide useful information for estimating body position, with comparable performance across the evaluated architectures.

\begin{table}[t]
  \centering
\caption{Performance of deep learning models for thermal image-based body-position classification. $\uparrow$ indicates higher is better.}
  \label{tab:pose}
  \small
  \setlength{\tabcolsep}{3pt}
  \resizebox{\textwidth}{!}{%
  \begin{tabular}{l c ccc ccc ccc}
    \toprule
    \multirow{2}{*}{\textbf{Model}} & \multirow{2}{*}{\textbf{Accuracy}$\uparrow$}
      & \multicolumn{3}{c}{\textbf{Supine}}
      & \multicolumn{3}{c}{\textbf{Left}}
      & \multicolumn{3}{c}{\textbf{Right}} \\
    \cmidrule(lr){3-5} \cmidrule(lr){6-8} \cmidrule(lr){9-11}
    & & \textbf{Precision}$\uparrow$ & \textbf{Recall}$\uparrow$ & \textbf{F1-Score}$\uparrow$
      & \textbf{Precision}$\uparrow$ & \textbf{Recall}$\uparrow$ & \textbf{F1-Score}$\uparrow$
      & \textbf{Precision}$\uparrow$ & \textbf{Recall}$\uparrow$ & \textbf{F1-Score}$\uparrow$ \\
    \midrule
    YOLOv8n-cls     & 0.937 & 0.973 & 0.946 & 0.960 & 0.842 & 0.928 & 0.883 & 0.912 & 0.946 & 0.929 \\
    YOLO11n-cls     & 0.936 & 0.967 & 0.950 & 0.959 & 0.844 & 0.904 & 0.873 & 0.918 & 0.947 & 0.932 \\
    ResNet18        & 0.945 & 0.972 & \textbf{0.956} & \textbf{0.964} & \textbf{0.861} & 0.931 & 0.894 & \textbf{0.930} & 0.952 & 0.941 \\
    ResNet50        & 0.942 & 0.970 & 0.954 & 0.962 & 0.852 & 0.918 & 0.884 & 0.922 & 0.954 & 0.938 \\
    EfficientNet-B0 & 0.941 & 0.968 & 0.954 & 0.961 & 0.855 & 0.913 & 0.883 & 0.919 & 0.953 & 0.936 \\
    DINOv2 ViT-S    & 0.936 & 0.970 & 0.944 & 0.957 & 0.835 & 0.918 & 0.875 & 0.907 & 0.957 & 0.931 \\
    MobileNetV3-L   & 0.937 & 0.969 & 0.949 & 0.959 & 0.847 & 0.919 & 0.882 & 0.914 & 0.943 & 0.928 \\
    ConvNeXt-T      & \textbf{0.947} & \textbf{0.975} & 0.954 & \textbf{0.964} & 0.858 & \textbf{0.940} & \textbf{0.897} & 0.925 & \textbf{0.962} & \textbf{0.943} \\
    \bottomrule
  \end{tabular}%
  }
\end{table}

\begin{table}[t]
  \centering
\caption{Patient-level classification of positional obstructive sleep apnea (POSA) versus non-positional OSA using nasal airflow (N) and thoracoabdominal motion (T), with and without SpO$_2$. POSA is defined following Cartwright's criterion as an AHI $\geq 5$ with a supine-to-non-supine AHI ratio $\geq 2$. ApneaEye uses signals and body positions estimated from thermal video, while the PSG-based comparison uses corresponding PSG-derived signals and reference annotations.}
  \label{tab:positional}
  \resizebox{\textwidth}{!}{%
  \begin{tabular}{l ccccc ccccc}
    \toprule
    \multirow{2}{*}{\makecell[l]{\textbf{Respiration}\\\textbf{channels}}}
      & \multicolumn{5}{c}{\textbf{ApneaEye}} & \multicolumn{5}{c}{\textbf{PSG}} \\
    \cmidrule(lr){2-6} \cmidrule(lr){7-11}
    & \textbf{Accuracy}$\uparrow$ & \textbf{Precision}$\uparrow$ & \textbf{Recall}$\uparrow$ & \textbf{F1-Score}$\uparrow$ & \textbf{Specificity}$\uparrow$ & \textbf{Accuracy}$\uparrow$ & \textbf{Precision}$\uparrow$ & \textbf{Recall} & \textbf{F1-Score}$\uparrow$ & \textbf{Specificity}$\uparrow$ \\
    \midrule
    N+T          & 74.2 & 65.8 & 87.3 & 75.0 & 63.8 & 79.0 & 71.0 & 89.1 & 79.0 & 71.0 \\
    N+T+SpO$_2$  & 77.4 & 71.4 & 81.8 & 76.3 & 73.9 & 81.5 & 75.8 & 85.5 & 80.3 & 78.3 \\
    \bottomrule
  \end{tabular}}
\end{table}

\subsection{Positional Obstructive Sleep Apnea (POSA)}

ApneaEye can also identify positional OSA by estimating body position from thermal video (Section~\ref{sec:pose}), helping guide positional therapy.
Following Cartwright~\cite{cartwright1984}, we defined positional OSA as an AHI $\geq 5$ with a supine AHI at least twice the non-supine AHI (left, right or prone). 
Supine and non-supine AHI were calculated separately by dividing the number of respiratory events by the sleep duration in each position. We included only nights with at least 1~h of sleep in both positions, resulting in 138 assessable nights out of 261. Of these, 124 had OSA: 55 positional and 69 non-positional.
The reference combined hospital-scored events, PSG sleep staging and the PSG position sensor.
For ApneaEye, all components were estimated independently from thermal video: respiratory events from nasal airflow and thoracoabdominal signals, sleep periods from video actigraphy, and body position from the thermal pose model. For comparison, we applied the same event detection model to PSG nasal pressure and respiratory effort belts, using PSG-derived sleep periods and body positions. ApneaEye classified 74.2\% of patients correctly, identifying 87.3\% of those with positional
OSA (F1 75.0\%), against 79.0\% accuracy and 89.1\% recall for the PSG-based system
(Table~\ref{tab:positional}). Most errors were non-positional patients labelled positional, typically those whose supine-to-non-supine AHI ratio lay close to the
2$\times$ threshold, where small differences in event counts change the classification. Adding SpO$_2$
improved accuracy to 77.4\% for ApneaEye and 81.5\% for PSG. 

\subsection{Effect of Environment in sensing Respiration}\label{sec:Env_effect}

\begin{figure}[t]
    \centering
    \includegraphics[width=0.95\linewidth]{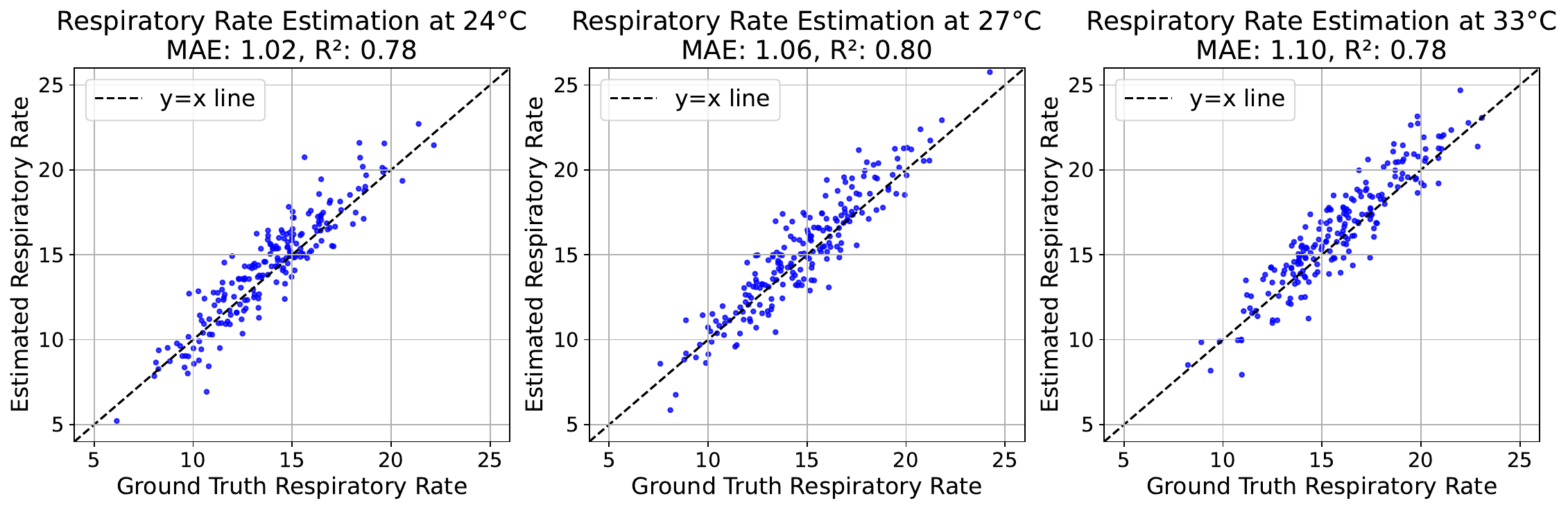}
    \caption{Scatter plots comparing estimated respiration rates from thermal video against ground truth chest belt measurements across three environmental temperatures: 24°C, 27°C, and 33°C. Each subplot corresponds to one temperature condition. The results demonstrate consistent estimation performance across varying temperatures.}
    \label{fig:Env-Factors}
\end{figure}

We conducted a controlled experiment to examine the effect of ambient temperature on respiration rate estimation from thermal video. Five healthy participants underwent one-hour nap sessions under three room-temperature conditions (24°C, 27°C, and 33°C), maintained using air conditioners. Reference respiration signals were simultaneously recorded using a chest belt, and respiration rates estimated from thermal video were compared against these measurements using 30-second windows with a 15-second overlap. The resulting mean absolute errors (MAE) were 1.02, 1.06, and 1.10 breaths per minute, with corresponding $R^2$ values of 0.78, 0.80, and 0.78 at 24°C, 27°C, and 33°C, respectively. The relatively small variation in MAE across the three conditions suggests comparable respiration estimation performance within the tested temperature range. Figure~\ref{fig:Env-Factors} presents scatter plots comparing estimated and reference respiration rates across the three conditions. While these results provide preliminary evidence of resilience to ambient temperature variations, the evaluation is limited to five healthy participants and short-duration nap recordings under controlled conditions, and does not establish robustness across broader environmental conditions or overnight recordings.

\section{Discussion and Limitations}

A central finding of ApneaEye is that the physiological information captured by a non-contact sensing system matters beyond its ability to detect respiratory motion. During obstructive apnea, respiratory effort may continue despite reduced or absent airflow, creating an ambiguity for systems that primarily measure thoracoabdominal movement. By recovering a thermal correlate of nasal airflow alongside respiratory motion, ApneaEye captures complementary aspects of breathing that are difficult to infer from motion alone. Although nostrils were visible for only 75.3\% of total sleep time, incorporating nasal respiration improved AHI estimation across all visibility groups, including nights with less than 50\% nostril visibility. This suggests that even intermittently available physiological information can improve downstream inference when combined with a more continuously observable signal such as thoracoabdominal motion. However, the AHI mean absolute error of 9.7 events/hour and relatively wide limits of agreement indicate that individual severity estimates remain uncertain, particularly near clinical decision thresholds. Incorporating SpO$_2$ reduced AHI error to 6.8 events/hour, highlighting the additional value of oxygenation information beyond thermal respiratory measurements. Together, these findings demonstrate the feasibility and value of recovering complementary physiological signals from a single non-contact sensor across extended overnight recordings, while highlighting the additional information needed for more accurate clinical screening.

Whether ApneaEye can support home-based sleep apnea screening remains an open question. Our clinical evaluation was conducted with controlled camera placement and a sleep-clinic population, and did not evaluate unsupervised setup or use in home environments. Additionally, central apnea events were substantially underrepresented in our dataset, with only 525 events compared to approximately 27,000 hypopnea and 16,000 obstructive apnea events. This class imbalance limits our ability to develop and reliably evaluate a classifier that distinguishes between different apnea event subtypes, particularly central and obstructive apnea. Future work will involve deploying ApneaEye in a broader population across multiple nights, with concurrent home sleep apnea testing as a reference, to assess sensitivity and specificity at clinically meaningful AHI thresholds. Such deployments must also evaluate performance across diverse body types, sleeping environments, and bed-partner configurations. A key prerequisite is an automated quality assurance mechanism that provides real-time feedback on camera placement, nasal visibility, and respiratory signal quality, enabling users to configure the system independently and identify periods of unreliable sensing.

Finally, the comparatively lower performance in hypopnea classification (65\%) highlights the difficulty of distinguishing partial airflow reductions using thermal respiratory signals alone. Clinical hypopnea scoring requires evidence of oxygen desaturation or cortical arousal in addition to reduced airflow, neither of which is directly measured by ApneaEye. The improvement observed when incorporating PSG-derived SpO$_2$ suggests that oxygenation measurements can provide valuable complementary information. Future work will investigate whether oximetry from commodity wearable devices can provide similar benefits, potentially enabling a hybrid screening approach that balances minimal instrumentation with improved clinical accuracy.

\section{Conclusion}

We presented ApneaEye, a non-contact sleep apnea monitoring system that uses a single commodity thermal camera to estimate nasal airflow-related dynamics and thoracoabdominal respiratory motion. Evaluated across 261 clinical overnight recordings totaling more than 2,000 hours, ApneaEye demonstrates the feasibility of recovering complementary respiratory information from thermal imagery under naturally varying sleeping conditions. Incorporating nasal airflow-related measurements improves AHI estimation and increases exact apnea severity classification accuracy by 8.7 percentage points compared with motion-only sensing. Our evaluation also characterizes practical challenges in extended thermal respiratory monitoring, including small nasal regions, motion artifacts, and intermittent nostril visibility. Together, these findings establish a foundation for future work on accessible, minimally intrusive sleep apnea monitoring, while highlighting the need for further validation in unsupervised home environments.


\clearpage



\bibliographystyle{ACM-Reference-Format}
\bibliography{sample-base}

@String{Computing = "Computing" }

@String{Computer = "{IEEE} Computer" }

@String{Springer = "Springer-Verlag" }

@article{chen2021apneadetector,
author = {Chen, Xianda and Xiao, Yifei and Tang, Yeming and Fernandez-Mendoza, Julio and Cao, Guohong},
title = {ApneaDetector: Detecting Sleep Apnea with Smartwatches},
year = {2021},
issue_date = {June 2021},
publisher = {Association for Computing Machinery},
address = {New York, NY, USA},
volume = {5},
number = {2},
url = {https://doi.org/10.1145/3463514},
doi = {10.1145/3463514},
journal = {Proc. ACM Interact. Mob. Wearable Ubiquitous Technol.},
month = jun,
articleno = {59},
numpages = {22}
}

@inproceedings{DeepApnea_Liu,
  title={Deepapnea: Deep learning based sleep apnea detection using smartwatches},
  author={Liu, Zida and Chen, Xianda and Ma, Fenglong and Fernandez-Mendoza, Julio and Cao, Guohong},
  booktitle={2024 IEEE International Conference on Pervasive Computing and Communications (PerCom)},
  pages={206--216},
  year={2024},
  organization={IEEE}
}

@inproceedings{nandakumar2015contactless,
author = {Nandakumar, Rajalakshmi and Gollakota, Shyamnath and Watson, Nathaniel},
title = {Contactless Sleep Apnea Detection on Smartphones},
year = {2015},
isbn = {9781450334945},
publisher = {Association for Computing Machinery},
address = {New York, NY, USA},
url = {https://doi.org/10.1145/2742647.2742674},
doi = {10.1145/2742647.2742674},
booktitle = {Proceedings of the 13th Annual International Conference on Mobile Systems, Applications, and Services},
pages = {45–57},
numpages = {13},
location = {Florence, Italy},
series = {MobiSys '15}
}

@article{korompili2021psg,
  title={PSG-Audio, a scored polysomnography dataset with simultaneous audio recordings for sleep apnea studies},
  author={Korompili, Georgia and Amfilochiou, Anastasia and Kokkalas, Lampros and Mitilineos, Stelios A and Tatlas, Nicolas-Alexander and Kouvaras, Marios and Kastanakis, Emmanouil and Maniou, Chrysoula and Potirakis, Stelios M},
  journal={Scientific data},
  volume={8},
  number={1},
  pages={197},
  year={2021},
  publisher={Nature Publishing Group UK London}
}

@article{kapur2017clinical,
  title={Clinical practice guideline for diagnostic testing for adult obstructive sleep apnea: an American Academy of Sleep Medicine clinical practice guideline},
  author={Kapur, Vishesh K and Auckley, Dennis H and Chowdhuri, Susmita and Kuhlmann, David C and Mehra, Reena and Ramar, Kannan and Harrod, Christopher G},
  journal={Journal of clinical sleep medicine},
  volume={13},
  number={3},
  pages={479--504},
  year={2017},
  publisher={American Academy of Sleep Medicine}
}

@article{scebba2020multispectral,
  title={Multispectral video fusion for non-contact monitoring of respiratory rate and apnea},
  author={Scebba, Gaetano and Da Poian, Giulia and Karlen, Walter},
  journal={IEEE Transactions on Biomedical Engineering},
  volume={68},
  number={1},
  pages={350--359},
  year={2020},
  publisher={IEEE}
}

@ARTICLE{DengComplexMachinery,
  author={Deng, Fei and Dong, Jianwu and Wang, Xiangyu and Fang, Ying and Liu, Yu and Yu, Zhaofei and Liu, Jing and Chen, Feng},
  journal={IEEE Transactions on Instrumentation and Measurement}, 
  title={Design and Implementation of a Noncontact Sleep Monitoring System Using Infrared Cameras and Motion Sensor}, 
  year={2018},
  volume={67},
  number={7},
  pages={1555-1563},
  doi={10.1109/TIM.2017.2779358}
}

@article{WanMultiUserAcoustic,
author = {Wan, Haoran and Shi, Shuyu and Cao, Wenyu and Wang, Wei and Chen, Guihai},
title = {Multi-User Room-Scale Respiration Tracking Using COTS Acoustic Devices},
year = {2023},
issue_date = {November 2023},
publisher = {Association for Computing Machinery},
address = {New York, NY, USA},
volume = {19},
number = {4},
issn = {1550-4859},
url = {https://doi.org/10.1145/3594220},
doi = {10.1145/3594220},
journal = {ACM Trans. Sen. Netw.},
month = {jun},
articleno = {85},
numpages = {28}
}

@article{GuoBreathBandWifi,
author = {Guo, Zhengxin and Yuan, Wenyang and Gui, Linqing and Sheng, Biyun and Xiao, Fu},
title = {BreatheBand: A Fine-grained and Robust Respiration Monitor System Using WiFi Signals},
year = {2023},
issue_date = {November 2023},
publisher = {Association for Computing Machinery},
address = {New York, NY, USA},
volume = {19},
number = {4},
issn = {1550-4859},
url = {https://doi.org/10.1145/3582079},
doi = {10.1145/3582079},
journal = {ACM Trans. Sen. Netw.},
month = {may},
articleno = {82},
numpages = {18}
}

@article{LiuWifiReflected,
author = {Liu, Jinyi and Zeng, Youwei and Gu, Tao and Wang, Leye and Zhang, Daqing},
title = {WiPhone: Smartphone-based Respiration Monitoring Using Ambient Reflected WiFi Signals},
year = {2021},
issue_date = {March 2021},
publisher = {Association for Computing Machinery},
address = {New York, NY, USA},
volume = {5},
number = {1},
url = {https://doi.org/10.1145/3448092},
doi = {10.1145/3448092},
journal = {Proc. ACM Interact. Mob. Wearable Ubiquitous Technol.},
month = {mar},
articleno = {23},
numpages = {19}
}

@inproceedings{LiuWifiReflected_OffTheShelf,
author = {Liu, Jian and Wang, Yan and Chen, Yingying and Yang, Jie and Chen, Xu and Cheng, Jerry},
title = {Tracking Vital Signs During Sleep Leveraging Off-the-shelf WiFi},
year = {2015},
isbn = {9781450334891},
publisher = {Association for Computing Machinery},
address = {New York, NY, USA},
url = {https://doi.org/10.1145/2746285.2746303},
doi = {10.1145/2746285.2746303},
booktitle = {Proceedings of the 16th ACM International Symposium on Mobile Ad Hoc Networking and Computing},
pages = {267–276},
numpages = {10},
location = {Hangzhou, China},
series = {MobiHoc '15}
}

@inproceedings{Abdel_UBIbreath,
author = {Abdelnasser, Heba and Harras, Khaled A. and Youssef, Moustafa},
title = {UbiBreathe: A Ubiquitous non-Invasive WiFi-based Breathing Estimator},
year = {2015},
isbn = {9781450334891},
publisher = {Association for Computing Machinery},
address = {New York, NY, USA},
url = {https://doi.org/10.1145/2746285.2755969},
doi = {10.1145/2746285.2755969},
booktitle = {Proceedings of the 16th ACM International Symposium on Mobile Ad Hoc Networking and Computing},
pages = {277–286},
numpages = {10},
location = {Hangzhou, China},
series = {MobiHoc '15}
}

@article{XieWifiCSIcirclemodel,
  author={Xie, Wangdong and Gan, Liangyu and Huang, Leilei and Shi, Chunqi and Liu, Boxiao and Wu, Chia-Hsin and Lee, Yueh-Ting and Chen, Jinghong and Zhang, Runxi},
  journal={IEEE Transactions on Biomedical Circuits and Systems}, 
  title={A Real-Time Respiration Monitoring System Using WiFi Sensing Based on the Concentric Circle Model}, 
  year={2023},
  volume={17},
  number={2},
  pages={157-168},
  doi={10.1109/TBCAS.2022.3229435}}

@article{SIA2017_OSAAWareness,
title = {Awareness and knowledge of obstructive sleep apnea among the general population},
journal = {Sleep Medicine},
volume = {36},
pages = {10-17},
year = {2017},
issn = {1389-9457},
doi = {https://doi.org/10.1016/j.sleep.2017.03.030},
url = {https://www.sciencedirect.com/science/article/pii/S1389945717301892},
author = {Ching-Hui Sia and Yueheng Hong and Linda W.L. Tan and Rob M. {van Dam} and Chi-Hang Lee and Adeline Tan},
}

@article{AdhikariJoulesEye,
author = {Adhikary, Rishiraj and Sadeh, Maite and Batra, Nipun and Goel, Mayank},
title = {JoulesEye: Energy Expenditure Estimation and Respiration Sensing from Thermal Imagery While Exercising},
year = {2024},
issue_date = {December 2023},
publisher = {Association for Computing Machinery},
address = {New York, NY, USA},
volume = {7},
number = {4},
url = {https://doi.org/10.1145/3631422},
doi = {10.1145/3631422},
journal = {Proc. ACM Interact. Mob. Wearable Ubiquitous Technol.},
month = {jan},
articleno = {146},
numpages = {29}
}

@article{KristiPortablePSG,
author = {Kristiansen, Stein and Nikolaidis, Konstantinos and Plagemann, Thomas and Goebel, Vera and Traaen, Gunn Marit and \O{}verland, Britt and Aaker\o{}y, Lars and Hunt, Tove-Elizabeth and Loennechen, Jan P\r{a}l and Steinshamn, Sigurd Loe and Bendz, Christina Holt and Anfinsen, Ole-Gunnar and Gullestad, Lars and Akre, Harriet},
title = {Machine Learning for Sleep Apnea Detection with Unattended Sleep Monitoring at Home},
year = {2021},
issue_date = {April 2021},
publisher = {Association for Computing Machinery},
address = {New York, NY, USA},
volume = {2},
number = {2},
url = {https://doi.org/10.1145/3433987},
doi = {10.1145/3433987},
journal = {ACM Trans. Comput. Healthcare},
month = {feb},
articleno = {14},
numpages = {25}
}

@INPROCEEDINGS{LydenSmartwatchSpO2,
  author={Lyden, Brendan and Dair, Zachary and O'Reilly, Ruairí},
  booktitle={2022 IEEE-EMBS International Conference on Biomedical and Health Informatics (BHI)}, 
  title={Classification of Sleep Apnea via SpO2 in a Simulated Smartwatch Environment}, 
  year={2022},
  volume={},
  number={},
  pages={1-5},
  doi={10.1109/BHI56158.2022.9926966}}

@INPROCEEDINGS{ArleneSmartwatchSp02_2,
  author={John, Arlene and Nundy, Koushik Kumar and Cardiff, Barry and John, Deepu},
  booktitle={2021 43rd Annual International Conference of the IEEE Engineering in Medicine and Biology Society (EMBC)}, 
  title={SomnNET: An SpO2 Based Deep Learning Network for Sleep Apnea Detection in Smartwatches}, 
  year={2021},
  volume={},
  number={},
  pages={1961-1964},
  doi={10.1109/EMBC46164.2021.9631037}
}

@Article{SunRespMonitorRGB,
AUTHOR = {Sun, Yue and Wang, Wenjin and Long, Xi and Meftah, Mohammed and Tan, Tao and Shan, Caifeng and Aarts, Ronald M. and de With, Peter H. N.},
TITLE = {Respiration Monitoring for Premature Neonates in NICU},
JOURNAL = {Applied Sciences},
VOLUME = {9},
YEAR = {2019},
NUMBER = {23},
ARTICLE-NUMBER = {5246},
URL = {https://www.mdpi.com/2076-3417/9/23/5246},
ISSN = {2076-3417},
DOI = {10.3390/app9235246}
}

@article{Jin_ApneaDetectSnoring,
author = {Hui Jin  and Li-Ang Lee  and Lijuan Song  and Yanmei Li  and Jianxin Peng  and Nanshan Zhong  and Hsueh-Yu Li  and Xiaowen Zhang },
title = {Acoustic Analysis of Snoring in the Diagnosis of Obstructive Sleep Apnea Syndrome: A Call for More Rigorous Studies},
journal = {Journal of Clinical Sleep Medicine},
volume = {11},
number = {07},
pages = {765-771},
year = {2015},
doi = {10.5664/jcsm.4856},
URL = {https://jcsm.aasm.org/doi/abs/10.5664/jcsm.4856},
eprint = {https://jcsm.aasm.org/doi/pdf/10.5664/jcsm.4856}

}

@article{DING_ApneaDetectSnoring_VGG19-LSTM,
title = {Automatically detecting apnea-hypopnea snoring signal based on VGG19+LSTM},
journal = {Biomedical Signal Processing and Control},
volume = {80},
pages = {104351},
year = {2023},
issn = {1746-8094},
doi = {https://doi.org/10.1016/j.bspc.2022.104351},
url = {https://www.sciencedirect.com/science/article/pii/S1746809422008059},
author = {Li Ding and Jianxin Peng and Lijuan Song and Xiaowen Zhang},
}

@article{berry2012aasm,
  title={Rules for scoring respiratory events in sleep: update of the 2007 AASM manual for the scoring of sleep and associated events: deliberations of the sleep apnea definitions task force of the American Academy of Sleep Medicine},
  author={Berry, Richard B and Budhiraja, Rohit and Gottlieb, Daniel J and Gozal, David and Iber, Conrad and Kapur, Vishesh K and Marcus, Carole L and Mehra, Reena and Parthasarathy, Sairam and Quan, Stuart F and others},
  journal={Journal of clinical sleep medicine},
  volume={8},
  number={5},
  pages={597--619},
  year={2012},
  publisher={American Academy of Sleep Medicine}
}

@article{berry2014aasm,
  title={The AASM manual for the scoring of sleep and associated events: rules, terminology and technical specifications. version 2.1.},
  author={Berry, Richard B},
  journal={Darien Illinois: American Academy of Sleep Medicine},
  year={2014}
}

@misc{berry2017aasm,
  title={AASM scoring manual updates for 2017 (version 2.4)},
  author={Berry, Richard B and Brooks, Rita and Gamaldo, Charlene and Harding, Susan M and Lloyd, Robin M and Quan, Stuart F and Troester, Matthew T and Vaughn, Bradley V},
  journal={Journal of clinical sleep medicine},
  volume={13},
  number={5},
  pages={665--666},
  year={2017},
  publisher={American Academy of Sleep Medicine}
}

@article{benjafield2019estimation,
  title={Estimation of the global prevalence and burden of obstructive sleep apnoea: a literature-based analysis},
  author={Benjafield, Adam V and Ayas, Najib T and Eastwood, Peter R and Heinzer, Raphael and Ip, Mary SM and Morrell, Mary J and Nunez, Carlos M and Patel, Sanjay R and Penzel, Thomas and P{\'e}pin, Jean-Louis and others},
  journal={The Lancet respiratory medicine},
  volume={7},
  number={8},
  pages={687--698},
  year={2019},
  publisher={Elsevier}
}

@article{Akbarian2020_JMIR_NIR,
  title={Distinguishing obstructive versus central apneas in infrared video of sleep using deep learning: Validation study},
  author={Akbarian, Sina and Montazeri Ghahjaverestan, Nasim and Yadollahi, Azadeh and Taati, Babak},
  journal={Journal of Medical Internet Research},
  volume={22},
  number={5},
  pages={e17252},
  year={2020},
  publisher={JMIR Publications Toronto, Canada}
}

@article{akbarian2021_JMIR_NIR,
  title={Noncontact sleep monitoring with infrared video data to estimate sleep apnea severity and distinguish between positional and nonpositional sleep apnea: Model development and experimental validation},
  author={Akbarian, Sina and Ghahjaverestan, Nasim Montazeri and Yadollahi, Azadeh and Taati, Babak},
  journal={Journal of Medical Internet Research},
  volume={23},
  number={11},
  pages={e26524},
  year={2021},
  publisher={JMIR Publications Toronto, Canada}
}

@inproceedings{choi2024poster,
  title={Poster: home-based, on-device non-invasive obstructive sleep apnea monitoring with infrared video},
  author={Choi, You Rim and Eo, Gyeongseon and Yoon, Wonhyuck and Lee, Hyojin and Jang, Haemin and Kim, Dong Yoon and Shin, Hyun-Woo and Kim, Hyung-Sin},
  booktitle={Proceedings of the 22nd Annual International Conference on Mobile Systems, Applications and Services},
  pages={708--709},
  year={2024}
}

@article{chiuNIR_2024video,
  title={Video-based contactless detection of sleep apnea with deep-learning model},
  author={Chiu, Li-Wen and Chou, Yang-Ren and Wu, Yi-Chiao and Chung, Meng-Liang and Wu, Bing-Fei and Chou, Kun-Ta},
  journal={IEEE Transactions on Instrumentation and Measurement},
  volume={73},
  pages={1--13},
  year={2024},
  publisher={IEEE}
}

@article{maurya2025contactless,
  title={Contactless apnea event detection using visible-thermal imaging},
  author={Maurya, Lalit and Zwiggelaar, Reyer and Chawla, Deepak and Mahapatra, Prasant},
  journal={Signal, Image and Video Processing},
  volume={19},
  number={5},
  pages={389},
  year={2025},
  publisher={Springer}
}

@article{RespError2,
  title={A deep learning-based camera approach for vital sign monitoring using thermography images for ICU patients},
  author={Lyra, Simon and Mayer, Leon and Ou, Liyang and Chen, David and Timms, Paddy and Tay, Andrew and Chan, Peter Y and Ganse, Bergita and Leonhardt, Steffen and Hoog Antink, Christoph},
  journal={Sensors},
  volume={21},
  number={4},
  pages={1495},
  year={2021},
  publisher={MDPI}
}

@article{RespError4,
  title={Analysis of the correlation between infrared thermal sequence images of nostril area and respiratory rate},
  author={Jian, Bo-Lin and Huang, Min-Wei and Lee, Shin-Hsiung and Yau, Her-Terng},
  journal={IET Image Processing},
  volume={14},
  number={13},
  pages={3089--3094},
  year={2020},
  publisher={Wiley Online Library}
}

@article{RespError5,
  title={Respiratory rate estimation from thermal video data using spatio-temporal deep learning},
  author={Mozafari, Mohsen and Law, Andrew J and Goubran, Rafik A and Green, James R},
  journal={Sensors},
  volume={24},
  number={19},
  pages={6386},
  year={2024},
  publisher={MDPI}
}

@article{hayano2025detection,
  title={Detection of sleep apnea using smartphone-embedded inertial measurement unit},
  author={Hayano, Junichiro and Takeshima, Masahiro and Imanishi, Aya and Ogasawara, Masaya and Yamada, Yasuko and Yuda, Emi and Mishima, Kazuo},
  journal={Scientific reports},
  volume={15},
  number={1},
  pages={14923},
  year={2025},
  publisher={Nature Publishing Group UK London}
}

@article{boiko2023accelerometer,
  title={Accelerometer based system for unobtrusive sleep apnea detection},
  author={Boiko, Andrei and Gaiduk, Maksym and Seepold, Ralf and Madrid, Natividad Mart{\'\i}nez},
  journal={Procedia computer science},
  volume={225},
  pages={1592--1600},
  year={2023},
  publisher={Elsevier}
}

@inproceedings{zhou2025accelerometer,
  title={Pioneering Contactless Sleep Apnea Detection: Harnessing the Power of Smartphones},
  author={Zhou, Guangxin and Li, Yelei and Yan, Haizhou and Zhao, Wei and Zhang, Yi and Zeng, Zijing},
  booktitle={2025 47th Annual International Conference of the IEEE Engineering in Medicine and Biology Society (EMBC)},
  pages={1--5},
  year={2025},
  organization={IEEE}
}

@article{PARISH2009563,
title = {Sleep-Related Problems in Common Medical Conditions},
journal = {Chest},
volume = {135},
number = {2},
pages = {563-572},
year = {2009},
issn = {0012-3692},
doi = {https://doi.org/10.1378/chest.08-0934},
url = {https://www.sciencedirect.com/science/article/pii/S0012369209601525},
author = {James M. Parish},
}

@article{song2026contactless,
  title={Contactless Sleep Apnea Detection with Bodyseismography},
  author={Song, Yingjian and Chen, Jiayu and Zeng, Zixuan and Zhang, Yida and Pitafi, Zaid and Das, Deepak Kumar and Phillips, Bradley G and Kwon, Younghoon and Healy, William J and Zhang, Xiang and others},
  journal={Proceedings of the ACM on Interactive, Mobile, Wearable and Ubiquitous Technologies},
  volume={10},
  number={2},
  pages={1--31},
  year={2026},
  publisher={ACM, New York, NY, USA}
}

@inproceedings{song2026attention,
  title={Attention Feature Fusion with Cluster Contrastive Learning for Snoring and Breath-Holding Detection Using Seismic Sensing},
  author={Song, Yingjian and Chen, Jiayu and Zeng, Zixuan and Zhang, Yida and Pitafi, Zaid Farooq and Phillips, Bradley G and Zhang, Xiang and Dou, Fei and Song, Wenzhan},
  booktitle={2026 IEEE International Conference on Pervasive Computing and Communications (PerCom)},
  pages={1--10},
  year={2026},
  organization={IEEE}
}

@article{choi2026non,
  title={Non-contact on-device detection of obstructive sleep apnea from infrared video},
  author={Choi, You Rim and Park, Dongik and Lee, Hyun Kyung and Bae, Suahn and Park, Sehyun and Kim, Dong-Kyu and Kim, Yong Min and Park, Chan-Soon and Kim, Hyung-Sin and Shin, Hyun-Woo},
  journal={Nature Communications},
  year={2026},
  publisher={Nature Publishing Group}
}

@ARTICLE{Charlton2018_BR_numbers,
  author={Charlton, Peter H. and Birrenkott, Drew A. and Bonnici, Timothy and Pimentel, Marco A. F. and Johnson, Alistair E. W. and Alastruey, Jordi and Tarassenko, Lionel and Watkinson, Peter J. and Beale, Richard and Clifton, David A.},
  journal={IEEE Reviews in Biomedical Engineering}, 
  title={Breathing Rate Estimation From the Electrocardiogram and Photoplethysmogram: A Review}, 
  year={2018},
  volume={11},
  number={},
  pages={2-20},
  doi={10.1109/RBME.2017.2763681}}

@article{keshvani2019_BR_number,
  title={Improving respiratory rate accuracy in the hospital: a quality improvement initiative},
  author={Keshvani, Neil and Berger, Kimberly and Gupta, Arjun and DePaola, Sheila and Nguyen, Oanh Kieu and Makam, Anil N},
  journal={Journal of hospital medicine},
  volume={14},
  number={11},
  pages={673--677},
  year={2019},
  publisher={Wiley Online Library}
}

@article{Shao2025PillowSense,
author = {Shao, Qijia and Chen, Junxiao and Leung, Ho Man Colman and Zhao, Meiqi and Xu, Ruoyu and Liu, Jiting and Garc{\'i}a, Lisa Maria DiSalvo and Jiang, Xiaofan and St-Onge, Marie-Pierre and Zhou, Xia},
title = {My Pillow Knows My Sleep: Sleep Monitoring with Computational Fabrics in the Pillowcase},
year = {2025},
issue_date = {September 2025},
publisher = {Association for Computing Machinery},
address = {New York, NY, USA},
volume = {9},
number = {3},
url = {https://doi.org/10.1145/3749550},
doi = {10.1145/3749550},
month = sep,
articleno = {128},
numpages = {26}
}

@article{Kazemi2025SmartwatchRR,
author = {Kazemi, Kianoosh and Azimi, Iman and Liljeberg, Pasi and Rahmani, Amir M.},
title = {Respiration Rate Estimation via Smartwatch-based Photoplethysmography and Accelerometer Data: A Transfer Learning Approach},
year = {2025},
issue_date = {March 2025},
publisher = {Association for Computing Machinery},
address = {New York, NY, USA},
volume = {9},
number = {1},
url = {https://doi.org/10.1145/3712280},
doi = {10.1145/3712280},
month = mar,
articleno = {7},
numpages = {24}
}

@inproceedings{liu2023contactless,
  title={Contactless respiratory rate monitoring for icu patients based on unsupervised learning},
  author={Liu, Zimeng and Huang, Bin and Lin, Chun-Liang and Wu, Chieh-Liang and Zhao, Changchen and Chao, Wen-Cheng and Wu, Yu-Cheng and Zheng, Yadan and Wang, Zhiru},
  booktitle={2023 IEEE/CVF Conference on Computer Vision and Pattern Recognition Workshops (CVPRW)},
  pages={6005--6014},
  year={2023},
  organization={IEEE}
}

@techreport{aast2021workforce,
  author      = {{American Association of Sleep Technologists}},
  title       = {{2021 Workforce Survey Report}},
  institution = {American Association of Sleep Technologists},
  year        = {2021},
  month       = aug,
  url         = {https://www.aastweb.org/Portals/0/Docs/News/AAST%202021%20Workforce%20Survey%20Report.pdf}
}

@article{garg2026sleep,
  title={Sleep testing in India: Are we missing the opportunity?—A commentary on the need for standardisation and innovation},
  author={Garg, Himanshu and Garg, Prerana},
  journal={Lung India: Official Organ of Indian Chest Society},
  volume={43},
  number={2},
  pages={113},
  year={2026}
}

@inproceedings{resnet,
  title={Deep residual learning for image recognition},
  author={He, Kaiming and Zhang, Xiangyu and Ren, Shaoqing and Sun, Jian},
  booktitle={Proceedings of the IEEE conference on computer vision and pattern recognition},
  pages={770--778},
  year={2016}
}

@inproceedings{efficientnet,
  title={Efficientnet: Rethinking model scaling for convolutional neural networks},
  author={Tan, Mingxing and Le, Quoc},
  booktitle={International conference on machine learning},
  pages={6105--6114},
  year={2019},
  organization={PmLR}
}

@inproceedings{mobilenetv3,
  title={Searching for mobilenetv3},
  author={Howard, Andrew and Sandler, Mark and Chu, Grace and Chen, Liang-Chieh and Chen, Bo and Tan, Mingxing and Wang, Weijun and Zhu, Yukun and Pang, Ruoming and Vasudevan, Vijay and others},
  booktitle={Proceedings of the IEEE/CVF international conference on computer vision},
  pages={1314--1324},
  year={2019}
}

@inproceedings{convnext,
  title={A convnet for the 2020s},
  author={Liu, Zhuang and Mao, Hanzi and Wu, Chao-Yuan and Feichtenhofer, Christoph and Darrell, Trevor and Xie, Saining},
  booktitle={2022 IEEE/CVF conference on computer vision and pattern recognition (CVPR)},
  pages={11966--11976},
  year={2022},
  organization={IEEE}
}

@inproceedings{yolov8,
  title={You only look once: Unified, real-time object detection},
  author={Redmon, Joseph and Divvala, Santosh and Girshick, Ross and Farhadi, Ali},
  booktitle={Proceedings of the IEEE conference on computer vision and pattern recognition},
  pages={779--788},
  year={2016}
}

@article{dinov2,
  title={Dinov2: Learning robust visual features without supervision},
  author={Oquab, Maxime and Darcet, Timoth{\'e}e and Moutakanni, Th{\'e}o and Vo, Huy and Szafraniec, Marc and Khalidov, Vasil and Fernandez, Pierre and Haziza, Daniel and Massa, Francisco and El-Nouby, Alaaeldin and others},
  journal={arXiv preprint arXiv:2304.07193},
  year={2023}
}

@article{cartwright1984,
  title={Effect of sleep position on sleep apnea severity.},
  author={Cartwright, Rosalind Dymond},
  journal={Sleep},
  volume={7},
  number={2},
  pages={110--114},
  year={1984}
}


\clearpage

\appendix

\section*{Appendix}

\section{Generative AI Disclosure}

Generative AI tools were used during manuscript preparation for grammatical correction, stylistic refinement, and rewriting of text. Claude Code was also used to assist with translating analysis ideas into code and generating figures from signals collected as part of our dataset. All AI-assisted outputs were reviewed and verified by the authors, who take full responsibility for the accuracy, integrity, and content of the manuscript.

\section{Tiny Object Detection Problem}\label{sec:tiny_object_problem}

\begin{figure}[h]
    \centering
    \includegraphics[width=1.0\linewidth]{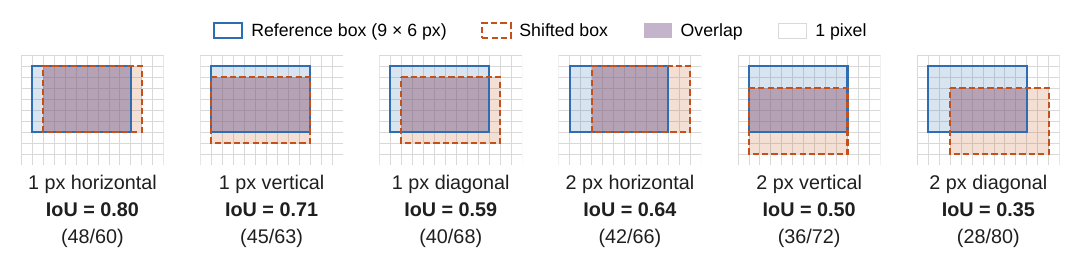}
\caption{Sensitivity of IoU to small localization errors for a
$9 \times 6$ pixel bounding box. Even a one-pixel displacement
reduces IoU to 0.80 (horizontal), 0.71 (vertical), and 0.59
(diagonal), illustrating the disproportionate effect of small
localization errors on tiny objects.}
    \label{fig:iou-sensitivity} 
\end{figure}

Intersection over Union (IoU) is a widely used metric for evaluating
bounding box localization. However, its sensitivity to localization
errors becomes particularly pronounced for tiny objects. When bounding
boxes occupy only a few pixels, even a small displacement of the
predicted box can substantially reduce the IoU, despite the prediction
being spatially close to the ground truth. For instance, as illustrated
in Fig.~\ref{fig:iou-sensitivity}, a reference bounding box of size
$9 \times 6$ pixels achieves an IoU of only 0.59 and 0.35 when shifted
by one and two pixels diagonally, respectively. Such reductions can
disproportionately penalize predictions that are reasonably accurate
in terms of their actual spatial localization. Consequently, IoU-based
evaluation alone may not adequately characterize localization
performance for tiny objects.

\end{document}